\documentclass{article}

\usepackage[final]{neurips_2021}
\setcitestyle{authoryear}
\usepackage{amsmath,amssymb,amsfonts}
\usepackage{algorithmic}
\usepackage{graphicx}
\usepackage{textcomp}
\usepackage{xcolor}
\usepackage{fancyhdr}
\usepackage[hyphens]{url}
\usepackage{wrapfig}
\usepackage{graphicx}
\def\BibTeX{{\rm B\kern-.05em{\sc i\kern-.025em b}\kern-.08em
    T\kern-.1667em\lower.7ex\hbox{E}\kern-.125emX}}

\usepackage{microtype}
\usepackage{markdown}
\usepackage{graphicx}
\usepackage{booktabs} 
\usepackage{subcaption}
\usepackage{xspace}

\usepackage{hyperref}
\usepackage{xcolor}

\definecolor{coolblack}{rgb}{0.0, 0.00, 0.50}

\extrafloats{100}

\usepackage{pifont}
\newcommand{\astar}{\textsc{astar}\xspace}
\newcommand{\bwaves}{\textsc{bwaves}\xspace}
\newcommand{\bzip}{\textsc{bzip}\xspace}
\newcommand{\cactusadm}{\textsc{cactusadm}\xspace}
\newcommand{\gems}{\textsc{gems}\xspace}
\newcommand{\lbm}{\textsc{lbm}\xspace}
\newcommand{\leslie}{\textsc{leslie3d}\xspace}
\newcommand{\libq}{\textsc{libq}\xspace}
\newcommand{\mcf}{\textsc{mcf}\xspace}
\newcommand{\milc}{\textsc{milc}\xspace}

\newcommand{\sphinx}{\textsc{sphinx3}\xspace}

\newcommand{\Belady}{Bélády\xspace}
\newcommand{\fakeparagraph}[1]{%
\vspace{0.25em}\textbf{#1}
}

\title{
Analyzing a Caching Model
}

\author{
\hspace{-0.5cm}
  Leon Sixt\thanks{%
  Work done during an internship at Google.
  }, \hspace{0.05cm}
Evan Zheran Liu,$^{\dagger,\circ}$
Marie Pellat,$^\circ$
James Wexler,$^\circ$
\\
\textbf{
Milad Hashemi,$^\circ$
Been Kim,$^\circ$%
Martin Maas$^\circ$
}
    \\
  $^*$ Freie Universität Berlin,
  $^\dagger$ Stanford University,
  $^\circ$ Google  \\
  Correspondence to: \texttt{mmaas@google.com}
}

\newcommand{\phasesFigSub}[2]{%
    \begin{subfigure}[t]{0.32\textwidth}
    \begin{center}
        \includegraphics[width=0.87\textwidth]{./figures/discovered_phases/#1.png}
    \end{center}
    \caption{#2}
    \label{fig:discovered_phases_#1}
    \end{subfigure}}
\newcommand{\FigPhases}{%
\begin{figure*}
\begin{center}
    \phasesFigSub{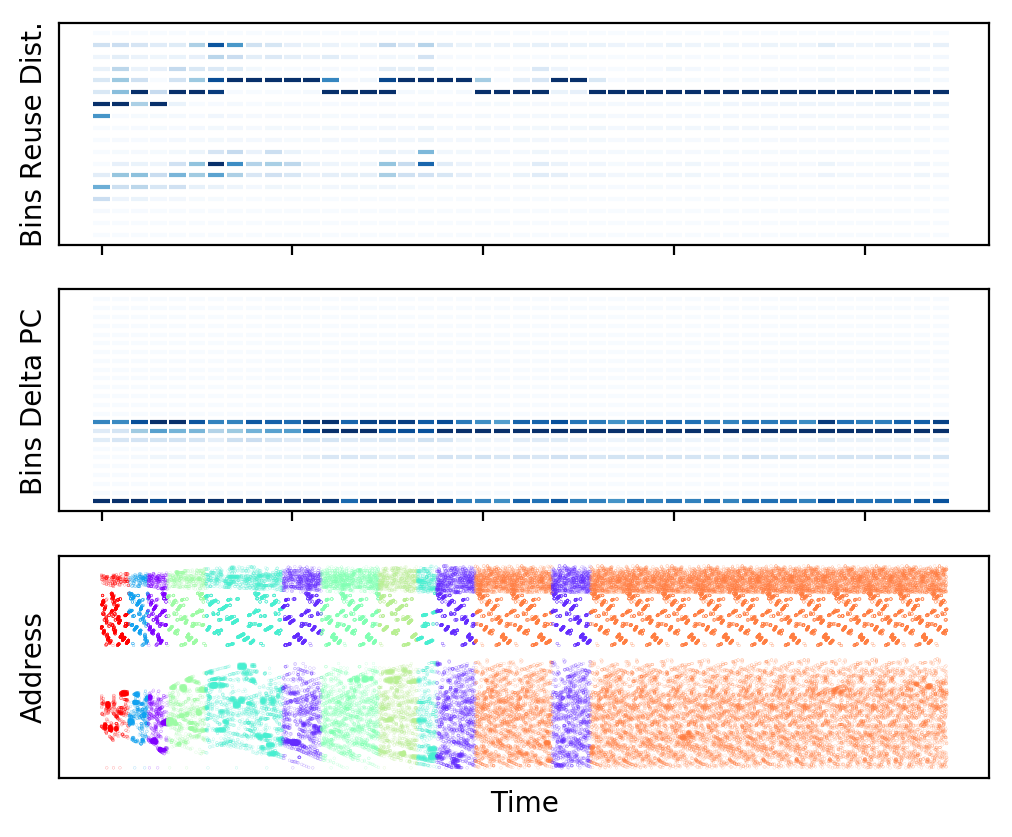}{\astar}
    \phasesFigSub{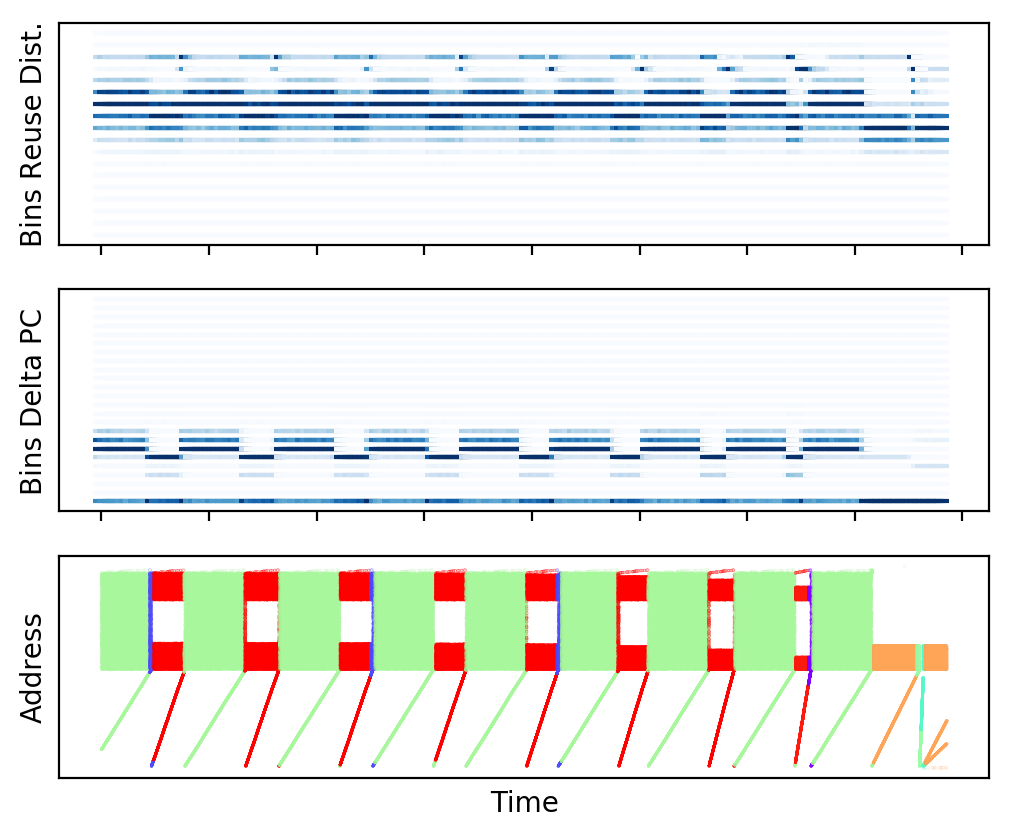}{\bzip}
    \phasesFigSub{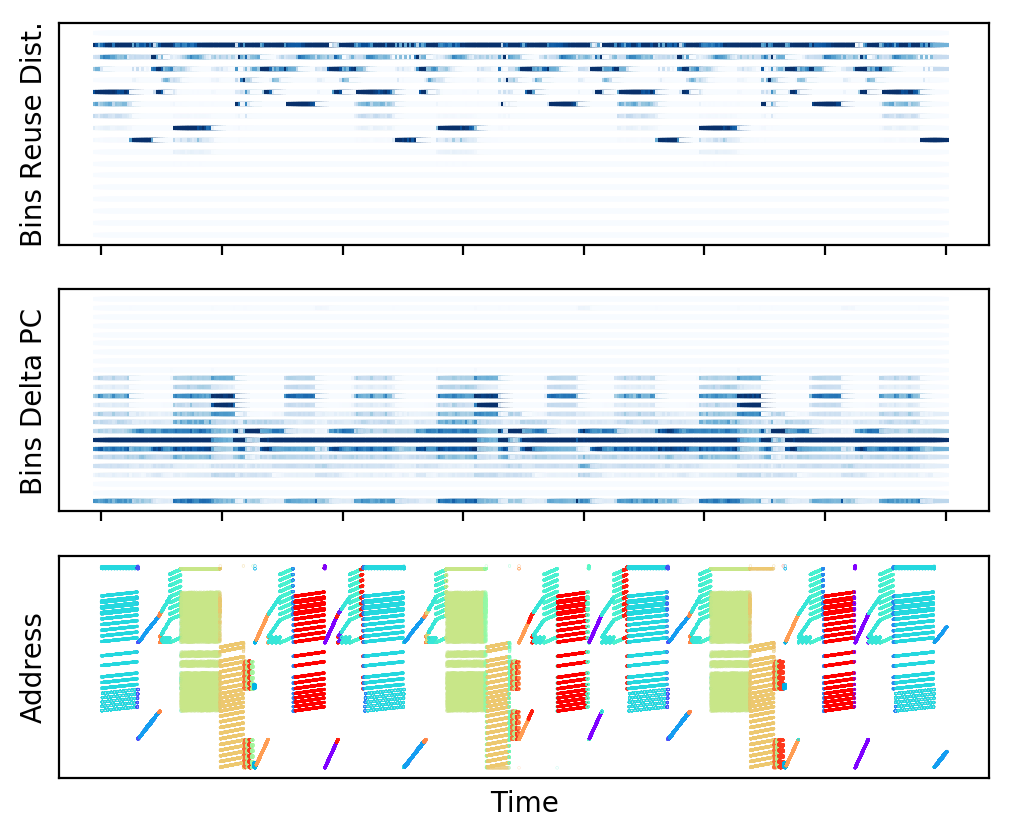}{\leslie}
\end{center}
\caption{Features used for finding the phases, and the phases themselves.
On top, the histograms of the Reuse Distance. Below, the histogram
of the delta program counter. On the bottom, the discovered phases.
}
\label{fig:phase_discovery}
\end{figure*}
}

\newcommand{\hitMissSubAppx}[2]{%
    \begin{subfigure}[t]{0.31\textwidth}
    \begin{center}
        \includegraphics[width=0.9\textwidth]{./figures/hits_miss/#1.png}
    \end{center}
    \caption{#2}
    \label{fig:hit_miss_appendix_#1}
    \end{subfigure}}

\newcommand{\FigAllHitMiss}{
\begin{figure*}
    \centering

    \hitMissSubAppx{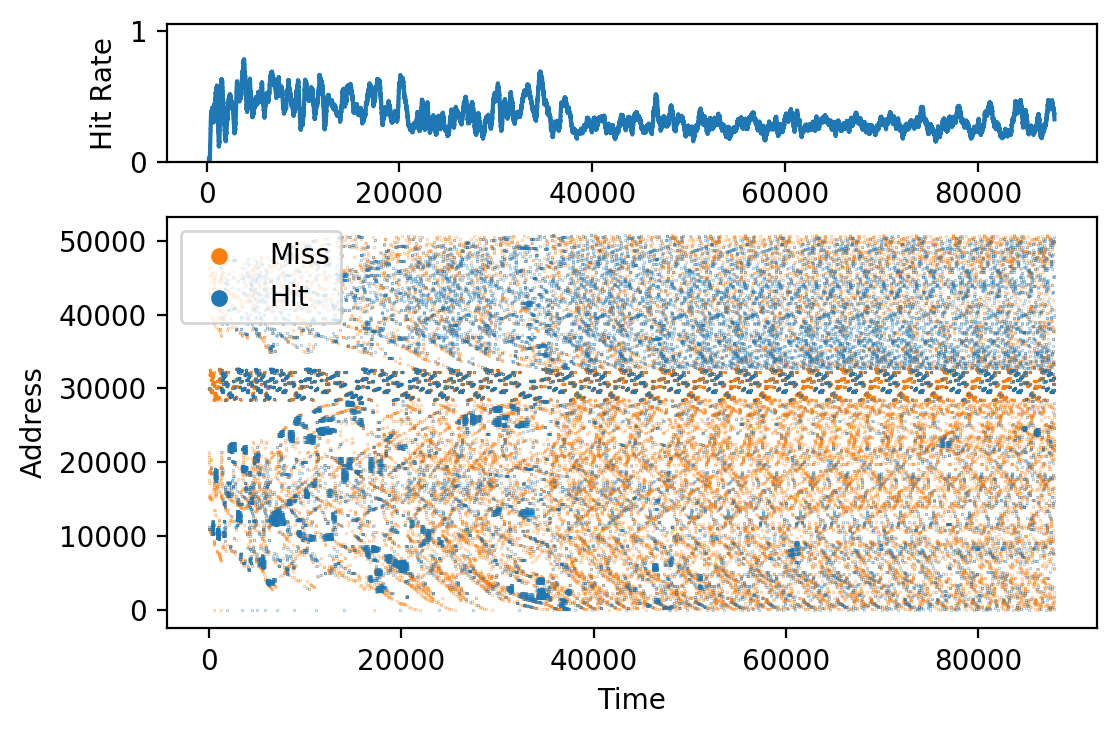}{\textsc{astar}}
    \hitMissSubAppx{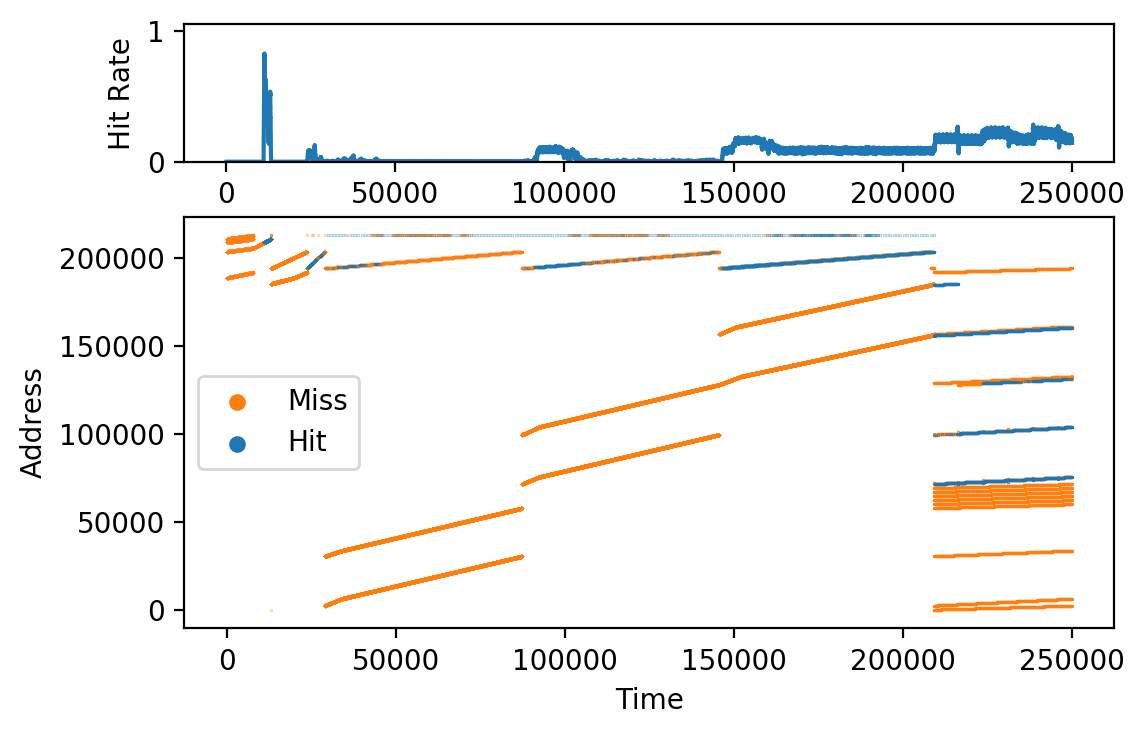}{\textsc{bwaves}}
    \hitMissSubAppx{bzip}{\textsc{bzip}}
    \hitMissSubAppx{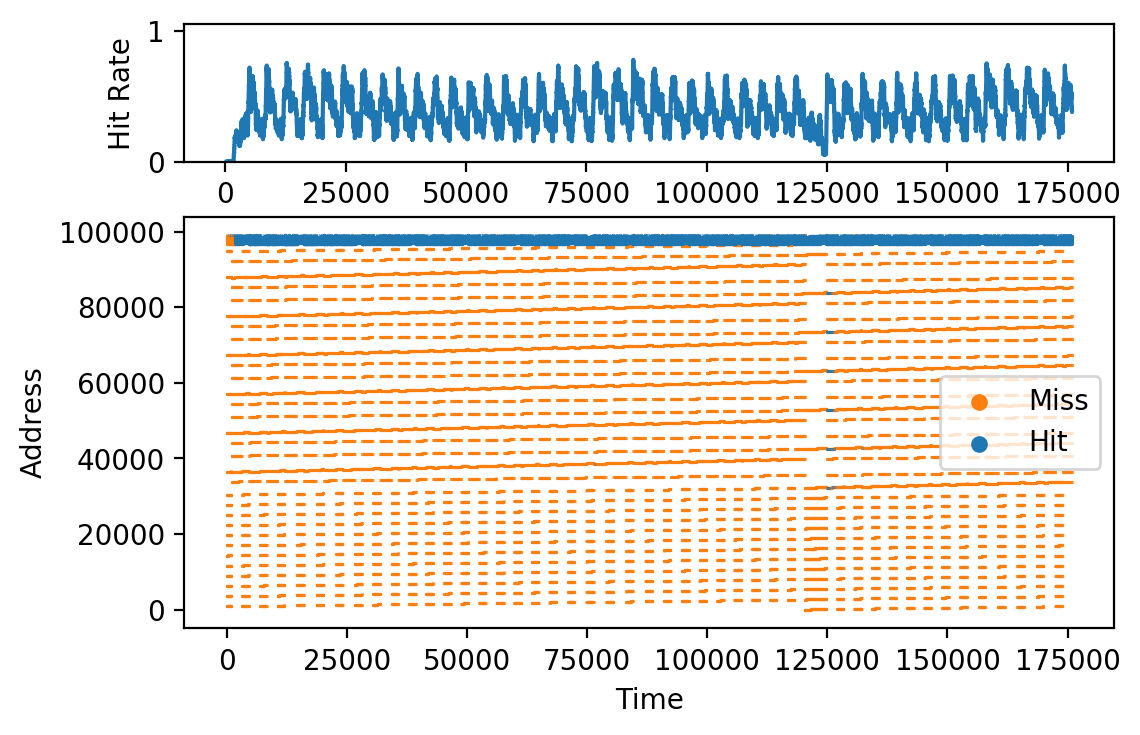}{\textsc{cactusadm}}
    \hitMissSubAppx{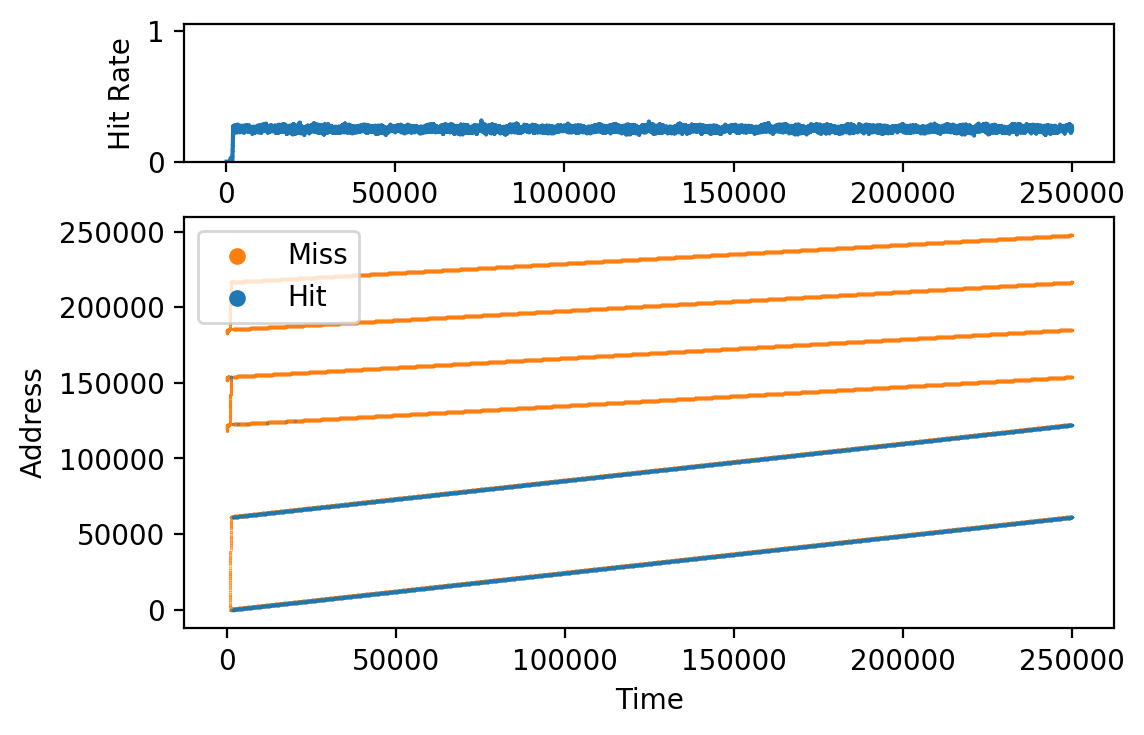}{\textsc{gems}}
    \hitMissSubAppx{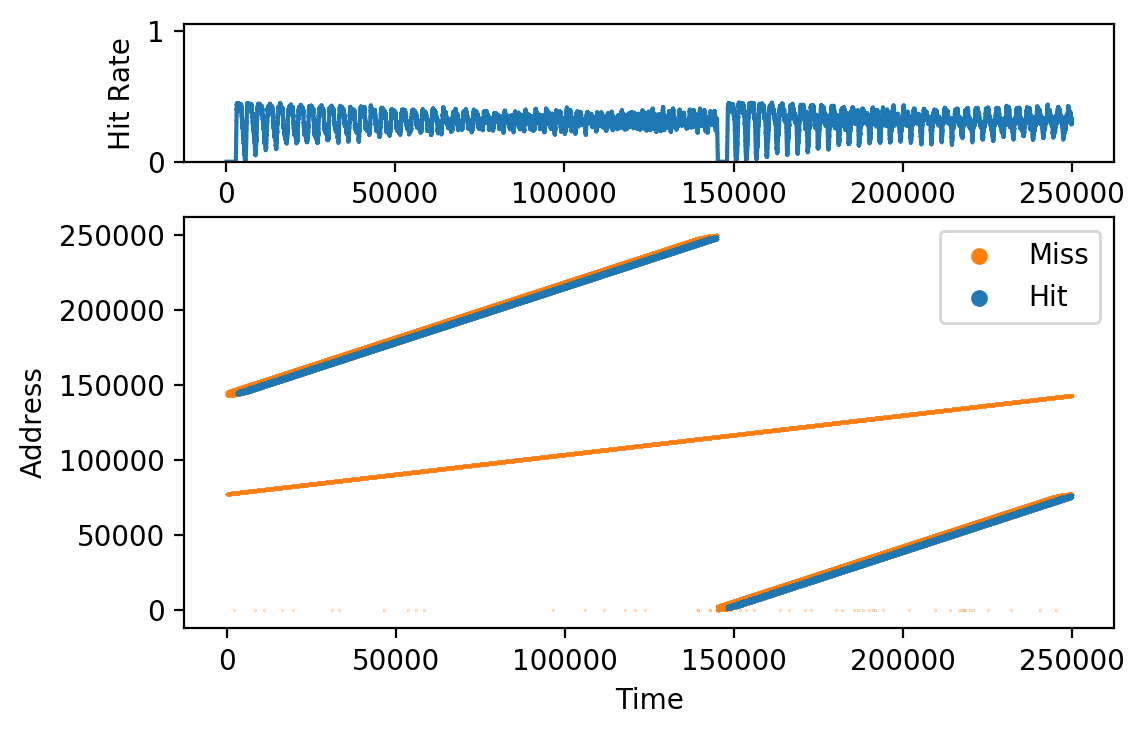}{\textsc{lbm}}
    \hitMissSubAppx{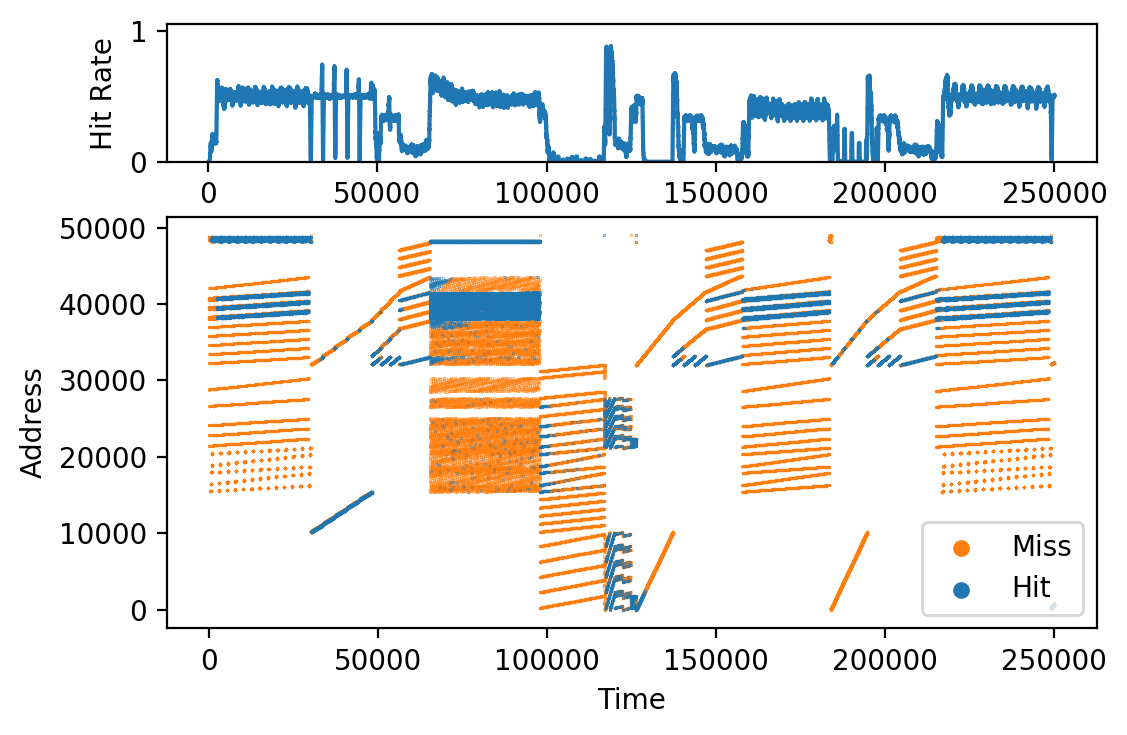}{\textsc{leslie3d}}
    \hitMissSubAppx{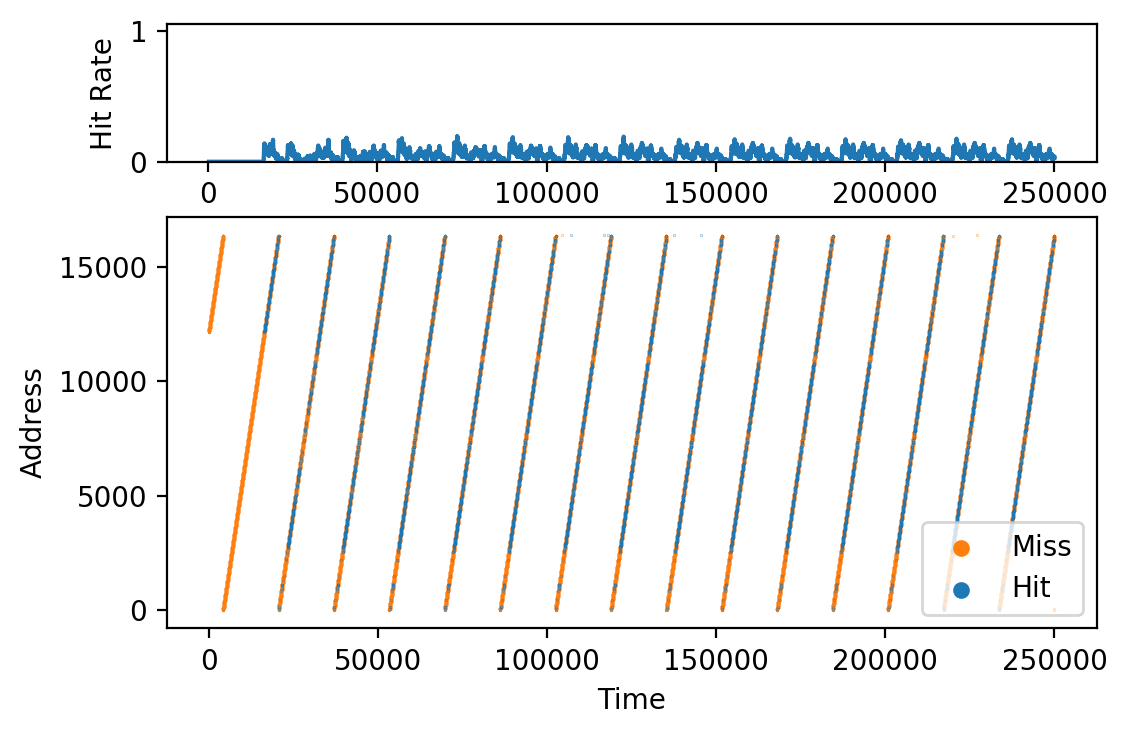}{\textsc{libq}}
    \hitMissSubAppx{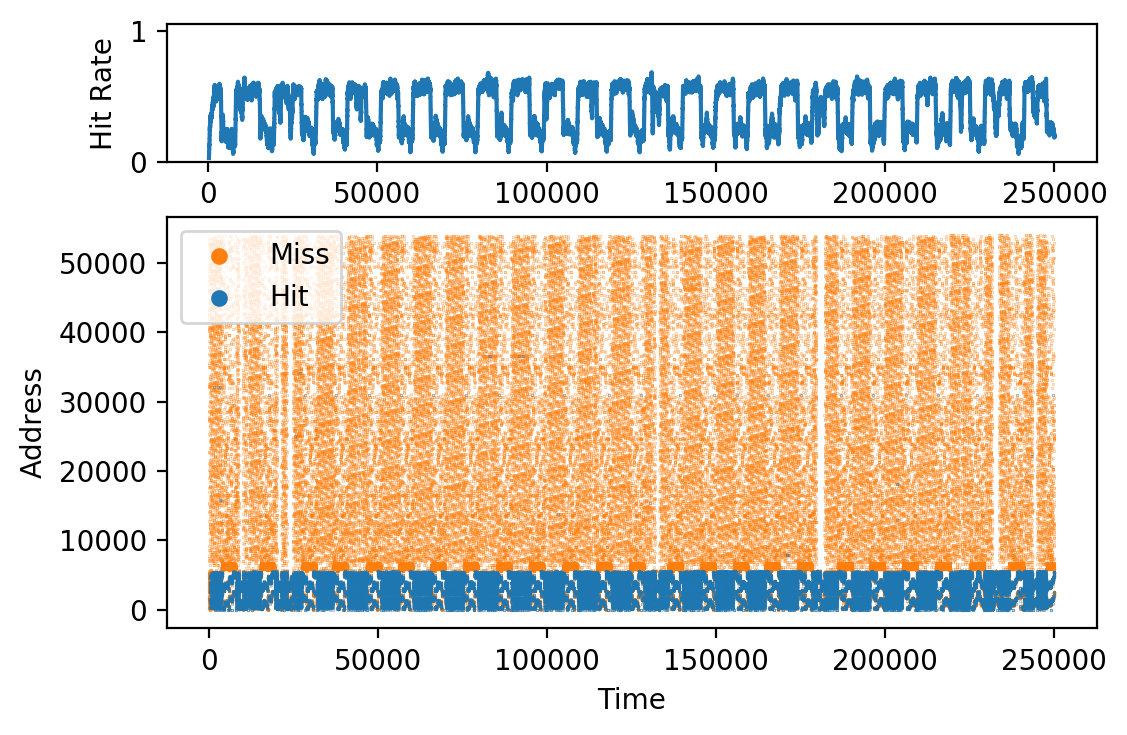}{\textsc{mcf}}
    \hitMissSubAppx{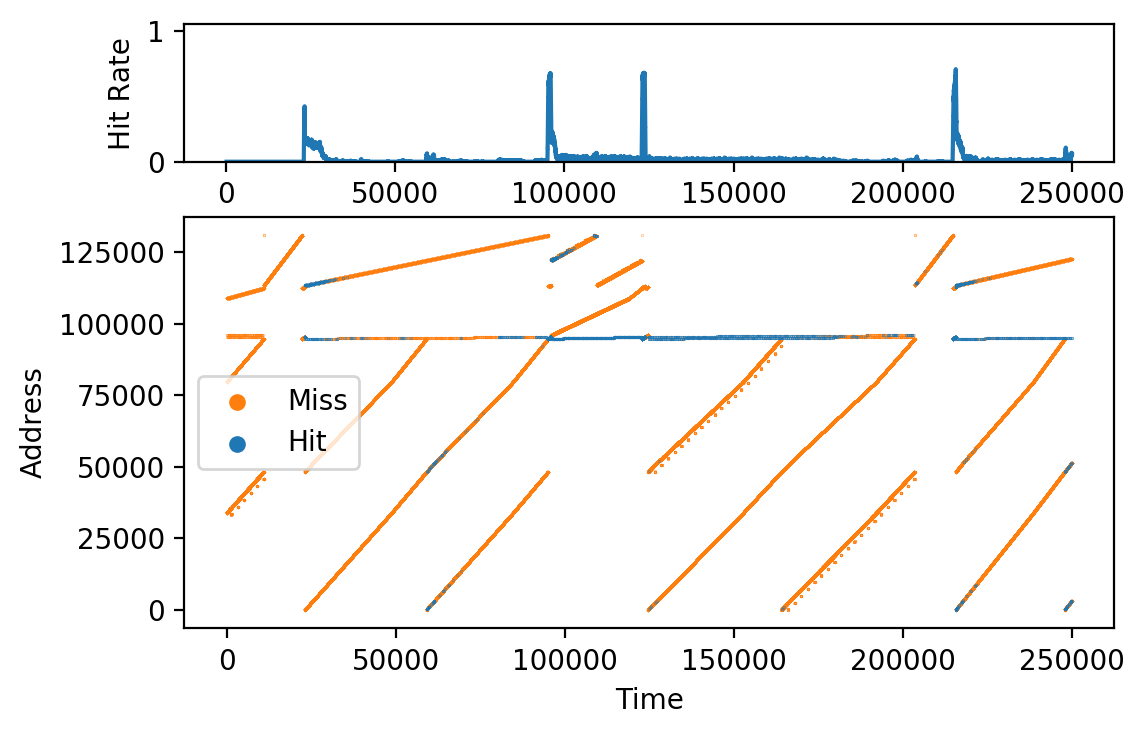}{\textsc{milc}}
    \hitMissSubAppx{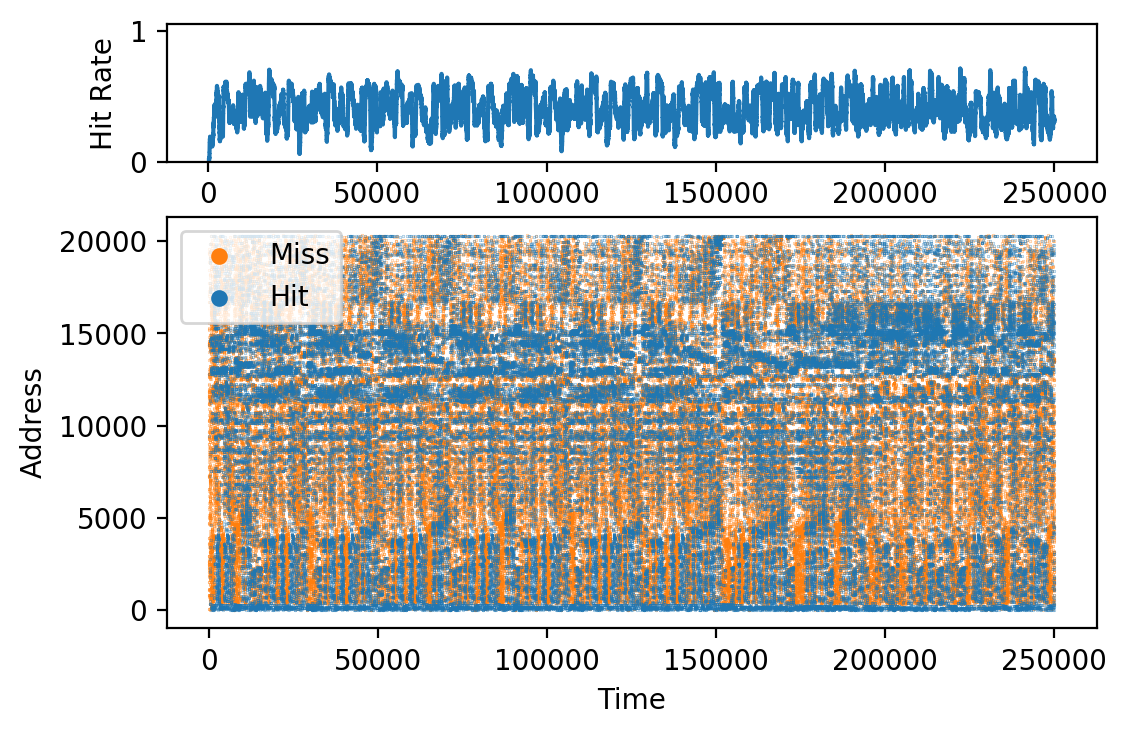}{\textsc{omnetpp}}
    \hitMissSubAppx{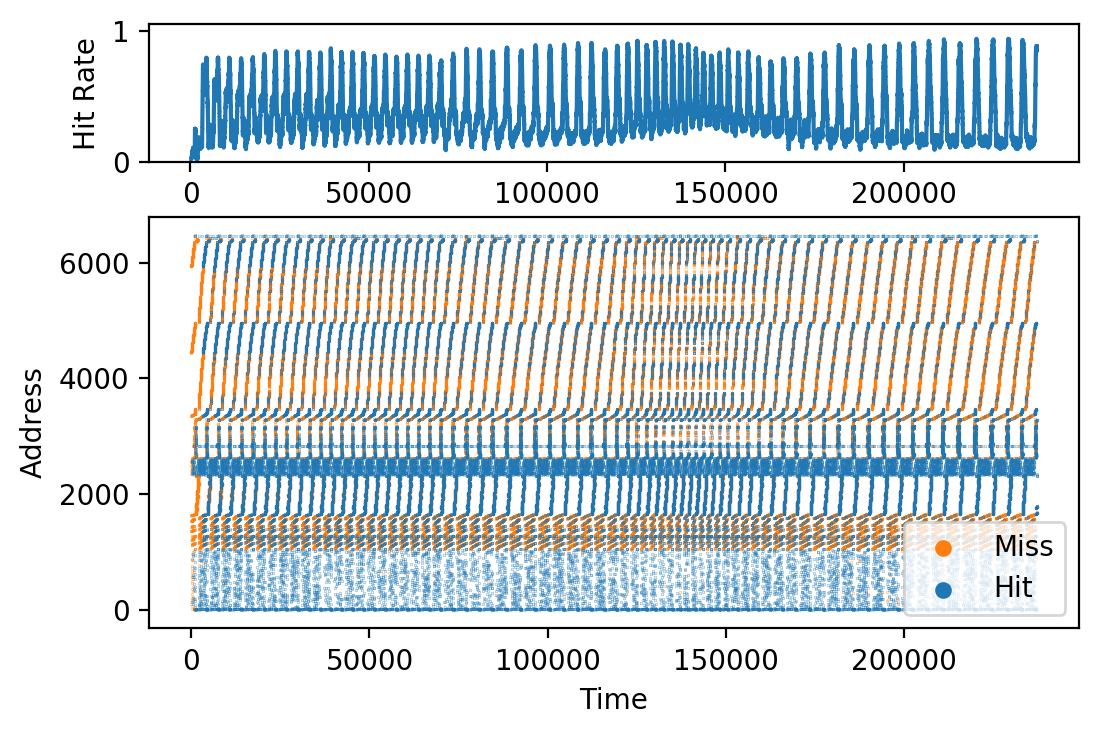}{\textsc{sphinx3}}
    \hitMissSubAppx{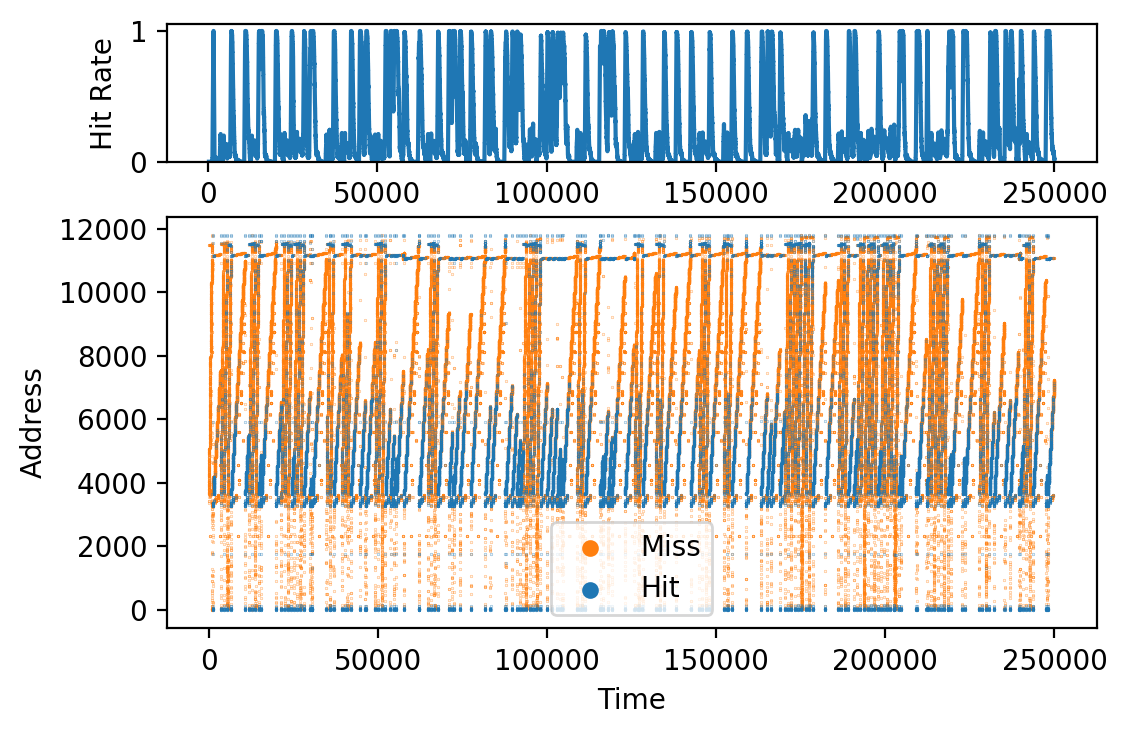}{\textsc{xalanc}}
    \caption{Cache hits and misses of the learned policy}
    \label{fig:hit_miss_appendix}
\end{figure*}
}

\newcommand{\pcaEmbeddingSubAppx}[2]{%
    \begin{subfigure}[t]{1.0\textwidth}
    \begin{center}
        \includegraphics[width=0.9\textwidth]{./figures/pca_embedding/address/#1.png}
    \end{center}
    \caption{#2}
    \label{fig:pca_address_appendix_#1}
    \end{subfigure}}

\newcommand{\FigPcaEmbeddingAppx}{%
\begin{figure*}
    \centering
    \pcaEmbeddingSubAppx{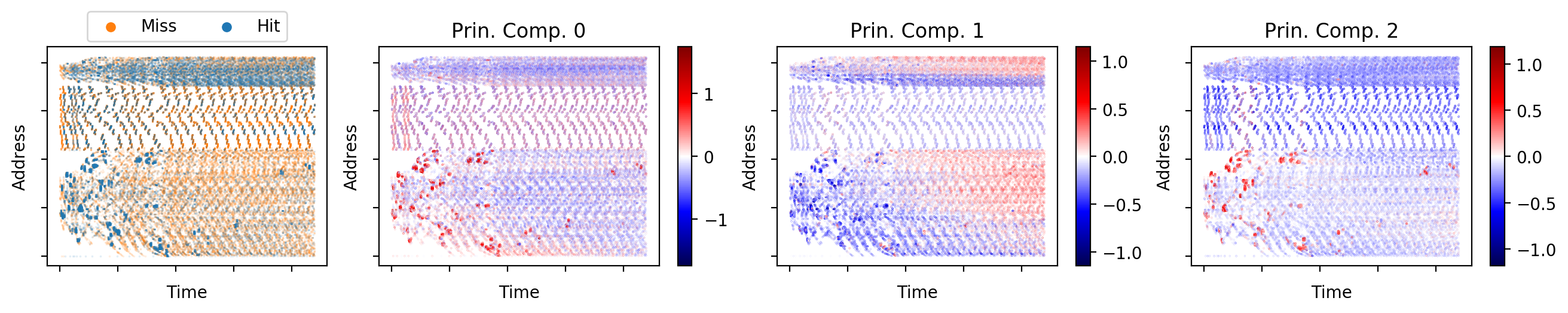}{\textsc{astar}}
    \pcaEmbeddingSubAppx{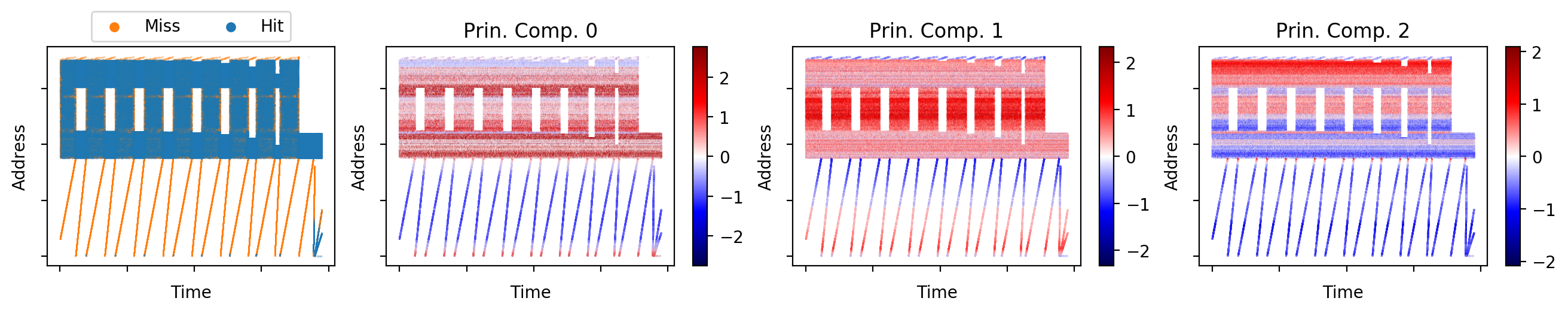}{\textsc{bzip}}
    \pcaEmbeddingSubAppx{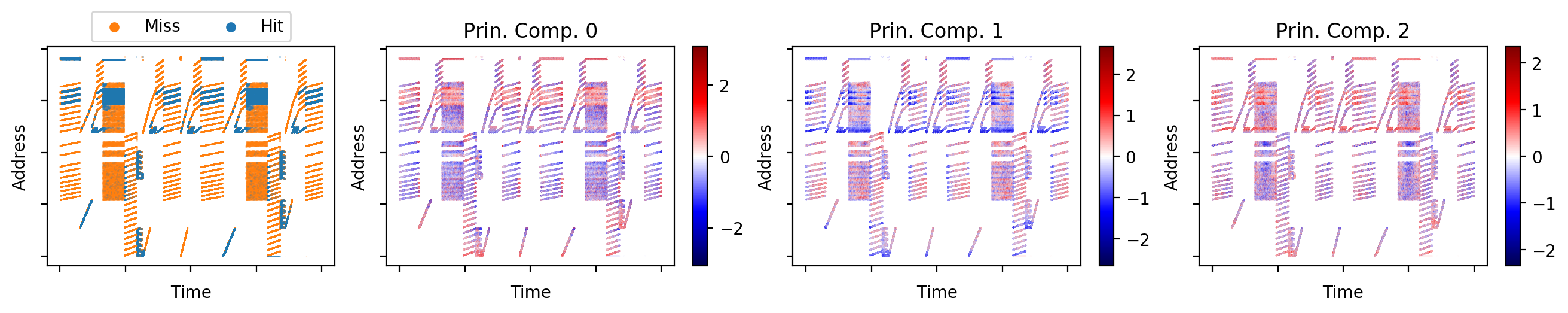}{\textsc{leslie3d}}
    \pcaEmbeddingSubAppx{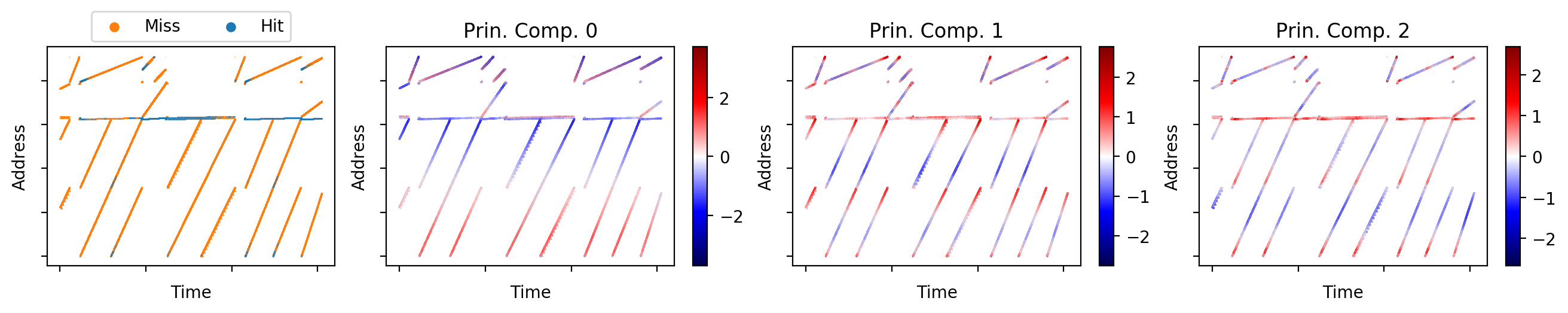}{\textsc{milc}}
    \pcaEmbeddingSubAppx{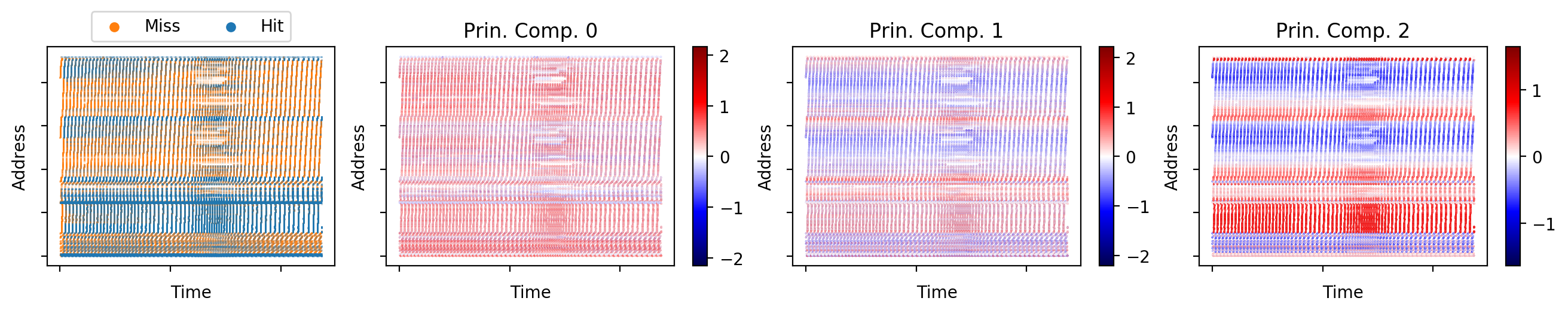}{\textsc{sphinx3}}
    \caption{PCA of address embeddings. First, the hit rates are shown, then the top-3
    principal components. The principal components (PC)
    correlate with the hits and misses.
    }
    \label{fig:pca_address_appendix}
\end{figure*}
}

\begin{document}
\maketitle


\begin{abstract}
Machine Learning has been successfully applied in systems
    applications such as memory prefetching and caching, where learned models have been shown to outperform heuristics. However, the lack of understanding
    the inner workings of these models -- interpretability -- remains a major
    obstacle for adoption in real-world deployments. Understanding
    a model's behavior can help system administrators and developers gain
    confidence in the model, understand risks, and debug unexpected behavior in production. Interpretability for models used in computer systems poses a particular
    challenge: Unlike ML models trained on images or text, the input domain (e.g.,
    memory access patterns, program counters) is not immediately interpretable. A major challenge is therefore to explain the model in terms of concepts that are approachable to a human practitioner.
    By analyzing a state-of-the-art caching model, we
    provide evidence that the model has learned concepts beyond simple
    statistics that can be leveraged for explanations.
    Our work provides a first step towards explanability of system ML models
     and highlights both promises and challenges of this emerging
    research area.
\end{abstract}

\section{Introduction}

In recent years, modern machine learning (ML) techniques have seen increasing
interest in the computer systems community \citep{maas2020_ml_taxonomy}, including black box
techniques such as neural networks. Research
has demonstrated that ML can be successfully applied to improve systems tasks
such as caching \citep{song2020,liu2020imitation,zhou2021}, job scheduling \citep{mao2019},
memory management \citep{maas2020,lujing2020}, and prefetching \citep{hashemi2018}.
The general strategy of these approaches is to replace a hard-coded \emph{system
policy}, such as a caching policy or a branch predictor, with a learned model.

Despite these promising research results, practical deployments of these
techniques are rare to find. One of the most common concerns about using ML in systems is the lack of \emph{interpretability}. For example, deploying
systems at scale requires a high degree of trust in their reliability --
for a web service, even a seemingly high 99.9\% reliability can mean that millions of people's lives are potentially affected by downtime.
To ensure reliability, the \emph{DevOps} engineers running
these systems need the ability to quickly introspect the inner
workings of a policy in case of failures, identify the input conditions that
prompted the policy to make a particular decision, and either rectify the policy
or those conditions. Traditional policies enable these steps by inspecting their
implementation, which black box ML models do not offer.
For this reason, past ML for systems projects have tried to use approaches such
as lookup tables or decision trees \citep{lifetime_nips_1996} that could be more interpretable, but are often less powerful then neural networks.

As a first step towards addressing this problem, we provide a quantitative case-study of adding interpretability to a state-of-the-art caching model by \cite{liu2020imitation}.
The model's task is to predict which memory addresses to keep in the cache and which to evict. The input is a sequence of memory addresses and program
counters recorded from different programs.
The representation of the model's input (low-level memory addresses and program counter) is difficult for humans to grasp, as programmers describe execution in high-level programming languages.
To provide approachable explanations, we need to find out whether
the model's operation  can be described in terms of comprehensible high-level concepts, or in terms of trace-specific low-level statistics (e.g. evict frequently accessed addresses less likely).
To this end, we investigate the generalization
and the abstraction level of the model's internal representation.
Our analysis shows that the model's embedding encodes trace-specific information, which may not be able to generalize to new traces and thus would be difficult to use in explanations.
However, we also find that the model can differentiate between different parts of the execution and that the model's activations represent distinct points in the execution, which points towards opportunities for using higher-level concepts to explain the behavior of the model.

\section{Setup: Caching Model and Analysis Goal}
\label{sec:setup}

\textbf{Caching} is an ubiquitous and difficult systems problem.
Caches are found across many systems, from processors to databases, operating systems, and content delivery
networks. As such, cache prediction models provide a good vector to study
interpretability for a wide array of scenarios.
The goal of a caching model is to learn a good cache replacement
policy, i.e., decide which item to evict when a new item is admitted to the cache.
\Belady's algorithm \citep{belady1966} is a well-known oracular policy, which evicts the data element used the farthest in the future.
While optimal, Bélády's policy is impractical, as future information is not available during online execution.
Therefore, many heuristics have been developed to approximate the
optimal replacement policy \citep{cachesurvey2001}. The most common is LRU -- evicting the Least Recently Used item.
We analyze the model from \citep{liu2020imitation} which uses imitation learning to
emulate the optimal policy.

\newcommand{\HsimpleCaching}{H1\xspace}
\newcommand{\HphaseVariablity}{H2\xspace}
\newcommand{\HstreamModifications}{H3\xspace}

\newcommand{\HpredicitabilityPhases}{H5\xspace}
\newcommand{\HaddressEmbedding}{H4\xspace}

\newcommand{\tableHitRates}{
\begin{table*}
    \caption{(\HsimpleCaching) Cache hit rates for different policies. \emph{\Belady} is the
    optimal policy. For the \emph{Address \& Phase} baseline, we used a lookup-table per address and phase and evict an address based on its frequency.  Entries
    where the baseline performs close to the model are shown in bold.
    }
    \small
    \centering
    \setlength{\tabcolsep}{3pt}
    \renewcommand{\arraystretch}{1.2}
    \begin{tabular}{l|ccccccccccccc}
        Policies  & astar & bwaves & bzip & cact. & gems & lbm  & leslie3d & libq & mcf  & milc & omne.  & sphinx3 & xalanc \\ \hline
        Belady & 43.5  & 8.7    & 78.4 & 38.8 & 26.5 & 31.3 & 31.9     & 5.8  & 46.8 & 2.4  & 45.1    & 38.2    & 33.3 \\
        Model  & 34.3  & 7.8    & 64.5 & 38.7      & 26.2 & {30.8} & {31.8}     & 5.1  & 41.1 & 2.1  & 40.3 & {36.6}    & {30.4}   \\
        Address \& Phase  & 27.7  & 1.3 & \textbf{62.1} & \textbf{35.1}      & 6.9  & 0.1  & 21.0     & 0.0 & \textbf{37.9} & 0.4  & 28.2    & \textbf{34.4}    & \textbf{22.4}   \\
    \end{tabular}
    \label{tab:phases}
    \vspace{-0.5cm}
\end{table*}
}

\begin{wrapfigure}{r}{5.0cm}
    \vspace{-0.45cm}
    \centering
    \includegraphics[width=\linewidth]{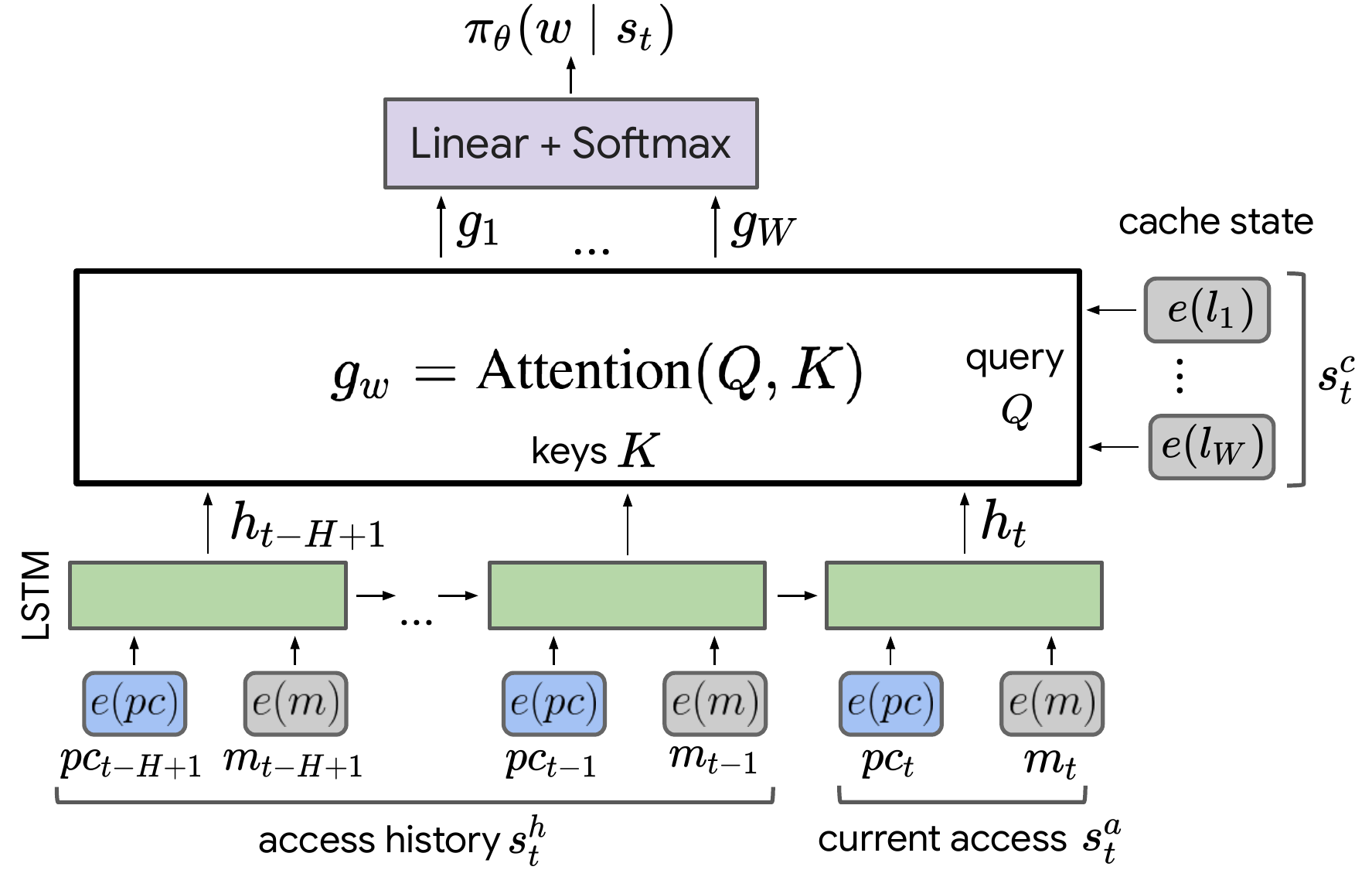}
    \caption{The analyzed caching model \citep{liu2020imitation}.
    }
    \label{fig:model}
\end{wrapfigure}
\textbf{The Caching Model}
Figure \ref{fig:model} shows an overview of the model's architecture.
The model has an LSTM layer, an attention layer, followed by a final fully connected layer.
The model's input is a time series of
cache accesses. Each access is encoded using an embedding of the program counter $e(p)$ and the memory address $e(m)$.
To make an eviction decision, the model calculates an eviction score.
First, the cached address embedding is used as attention key and
the LSTM's hidden state as attention query and value.
Based on the attention output, a fully
connected layer then computes the score.
The item with the highest score is evicted.
The model is trained using imitation learning to replicate the optimal policy during training, based on memory traces from the 
SPEC2006 benchmarks~\citep{henning2006spec}
The work trains a separate model for each execution trace (see
\citep{liu2020imitation} for details).

\section{Analyzing the Caching Model}
\label{sec:analyzing}
%
%
%
%


Explaining and characterizing the model's abstraction level can
be done in different ways. As a start, we analyze the
memory address trace and the model's performance statistics.

\newcommand{\hitMissFigBzip}{
\begin{wrapfigure}{r}{5.5cm}
    \vspace{-1cm}
    \begin{center}
        \includegraphics[width=\linewidth]{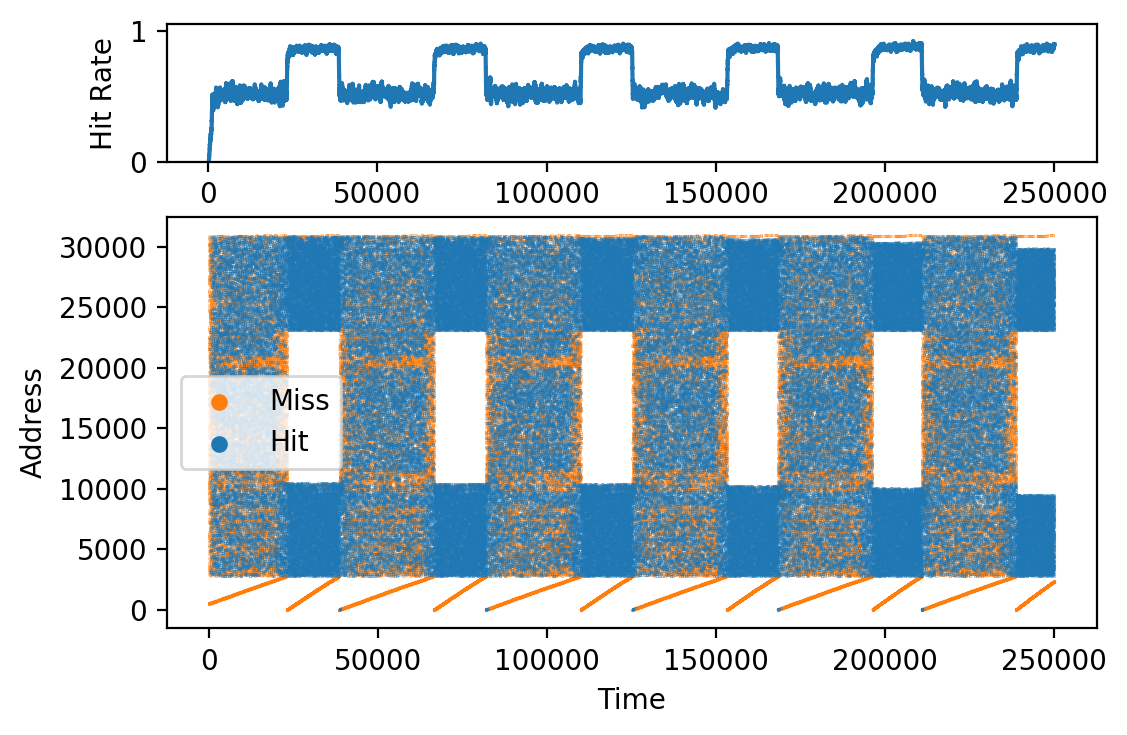}
    \end{center}
    \caption{Hits and misses.
    Per time step, we plot the access memory address and if it was
    a hit or misses.
    on the top, the hit rate of the model is shown.  }
    \label{fig:hit_miss}
    \vspace{-0.5cm}
\end{wrapfigure}
}
\fakeparagraph{Visualising the Execution and Performance Statistics}
For each time step, we plot the accessed address and whether it was a hit or a miss. The hit rate can be visualised using a rolling mean (see Figure \ref{fig:hit_miss} and Appendix Figure
\ref{fig:hit_miss_appendix}).
Inspecting these visual diagrams provides a first impression about the behaviour of the model on the different traces.
For example, the hit rate
is almost constant for
\lbm and \gems. The executions for
\bzip and \leslie are separated into repeating phases.
High frequency oscillations happen on \cactusadm, \sphinx, and
\mcf. \bwaves, \libq, \milc are difficult to cache and their the hit rates are zero most of the time.


\tableHitRates

\fakeparagraph{Information encoded in the address embedding}
In a further analysis, we look at possible limitations of the network architecture.
Here, the address embedding
stands out as it might limit generalization. Each memory address is assigned a unique embedding vector.
However, the memory layout can be very different
between two runs of the identical program (modern OSs employ address space randomization). We would therefore like to test which information is encoded in the memory address embedding.

\hitMissFigBzip
\emph{Experiment}: We apply Principal Component Analysis (PCA) on the address embeddings.
The first several Principal Components (PCs) account for the most variance and
are thus a good proxy for the encoded information in the embedding.

\emph{Result}:
In Figure \ref{fig:pca_address_appendix}, we display the hits and misses along with the
first three principal components of the address embeddings.  We see that trace-specific
features are encoded in the address embedding. For example, PC2
(principal component 2) of \bzip changes in tandem with distinct sequential access patterns in the lower address space.
The results from the other benchmarks also support the claim that the address
embeddings encode trace-specific features.
For example, for \astar the PC0 and PC2 seem to encode addresses that
lead to cache hits.
The trace-specific embedding means that a model cannot
generalize to new traces if the memory is allocated at a different location.

\newcommand{\FigModificationsBzip}{
\begin{figure*}[t]
    \vspace{-0.5cm}
    \def\bzipWidth{0.325\textwidth}
        \begin{subfigure}[b]{\bzipWidth}
            \centering
            \includegraphics[width=\linewidth]{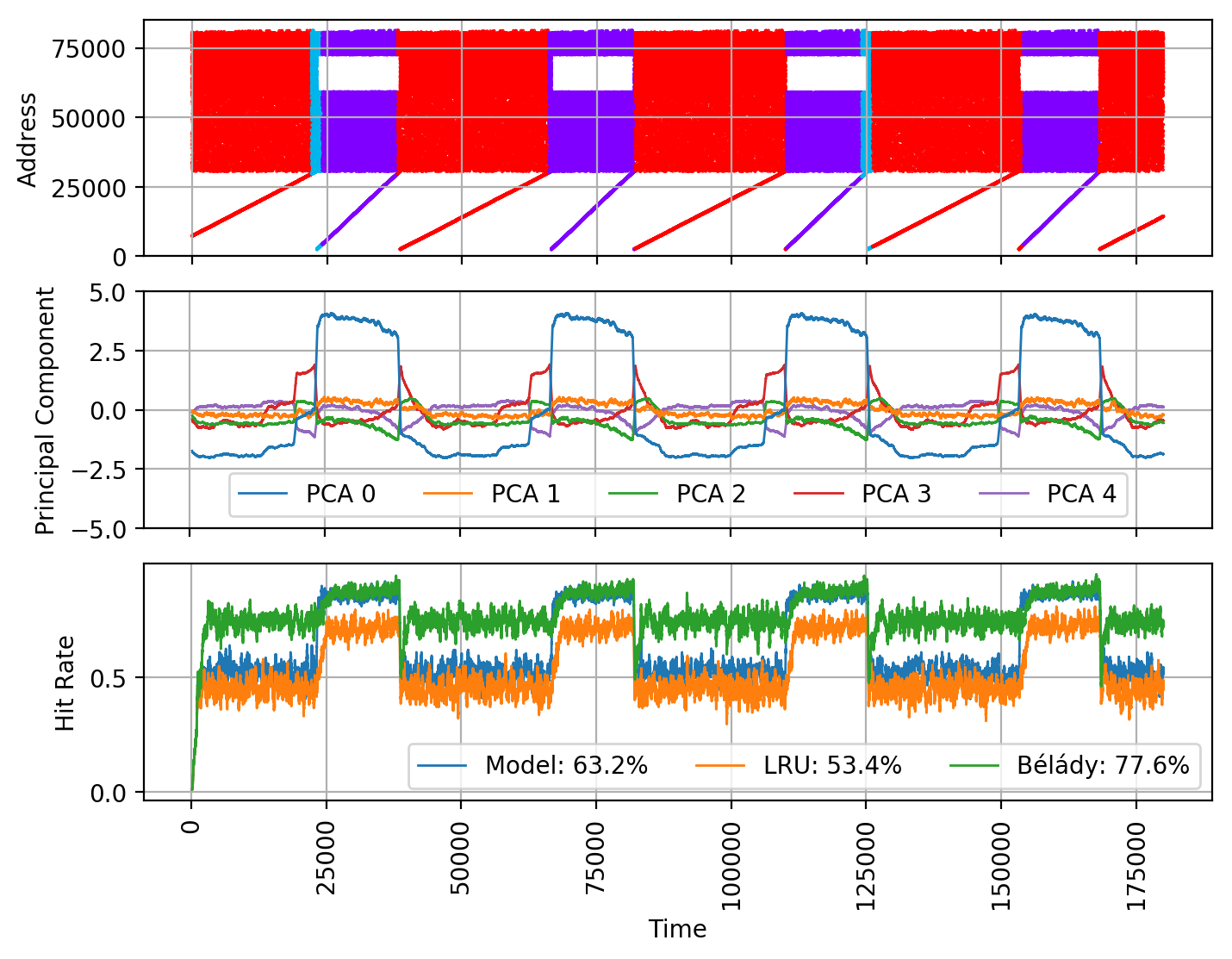}
            \caption{Original \textsc{bzip} memory trace}
            \label{fig:pca_bzip_original}
        \end{subfigure}
        \begin{subfigure}[b]{\bzipWidth}
            \centering
            \includegraphics[width=\linewidth]{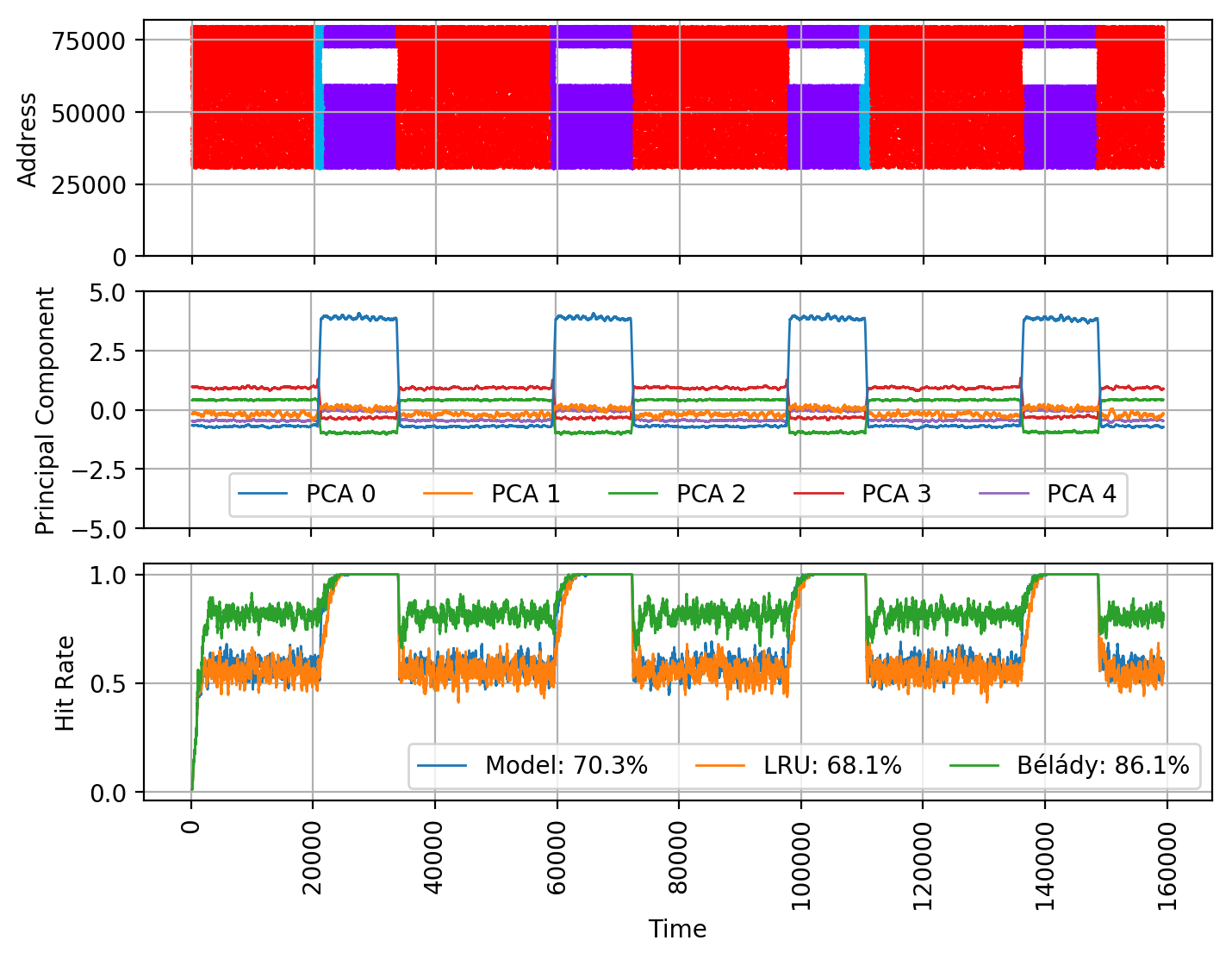}
            \caption{Lower streams removed} \label{fig:pca_bzip_removed_lower}
        \end{subfigure}
        \begin{subfigure}[b]{\bzipWidth}
            \centering
            \includegraphics[width=\linewidth]{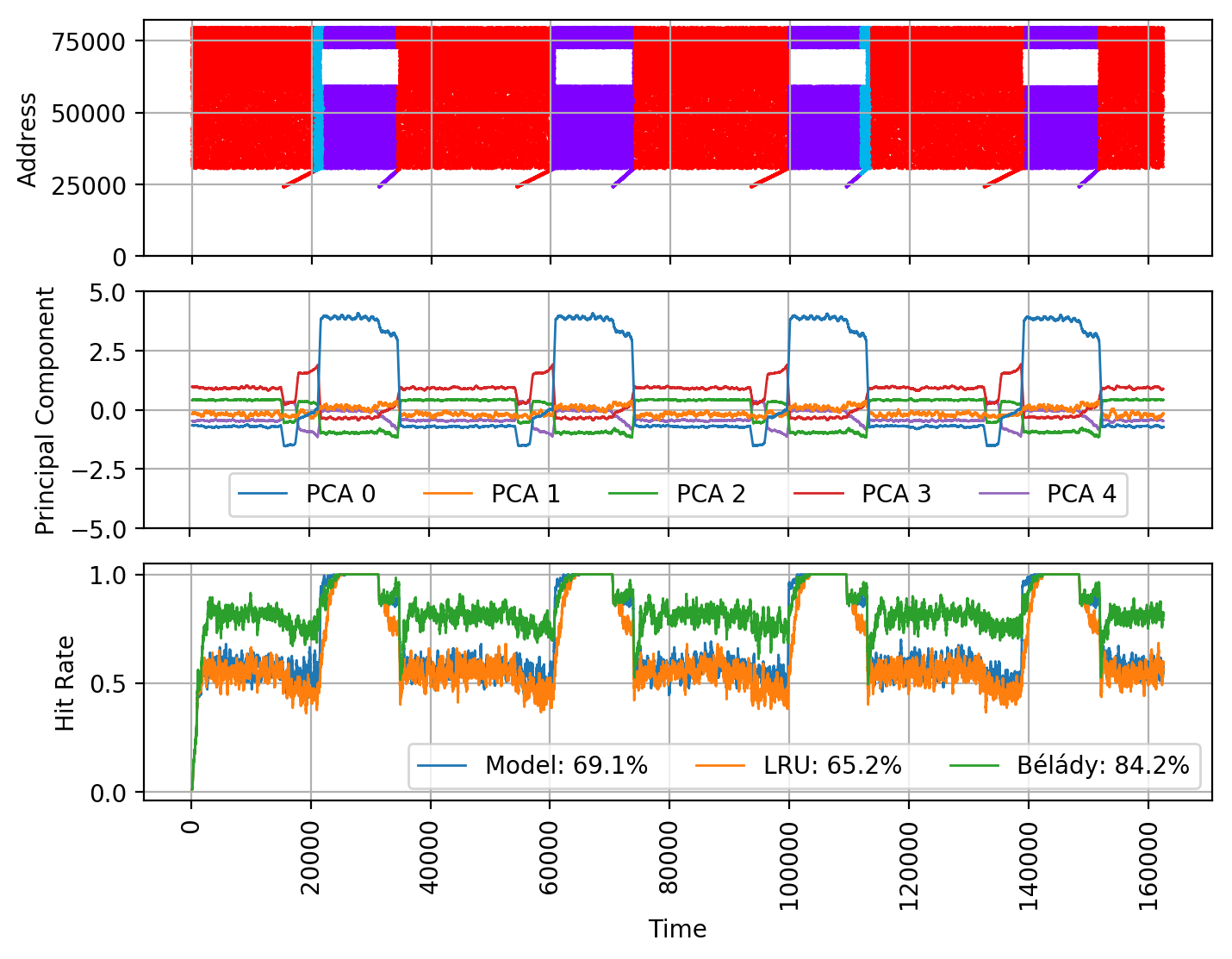}
            \caption{End part of lower stream included} \label{fig:pca_bzip_end_removed}
        \end{subfigure}
    \caption{
        On top, the memory trace of \textsc{bzip} with sequential access patterns (streams) at the lower end of the address space.
        In the middle, the top-5 principal components.
        On the bottom, the hit rates of the model, LRU and \Belady.
        \textbf{(a)} the original trace. A distinct jump
        in PCA componenents can be seen at at end of the red phase.
    \textbf{(b)} when the streams are removed, all variability with a phase disappears -- showing that the streams are responsible for encoding differences within a phase.
    \textbf{(c)} Including only the end part of the lower stream
    still shows the distinct jump -- suggesting that the model's behavior
    depends on specific memory addresses.
}
\label{fig:pca_bzip}
\vspace{-0.5cm}
\end{figure*}
}

\newcommand{\FigModificationsMilc}{
\begin{figure*}
    \begin{subfigure}{0.33\textwidth} \centering
    \includegraphics[width=0.85\linewidth]{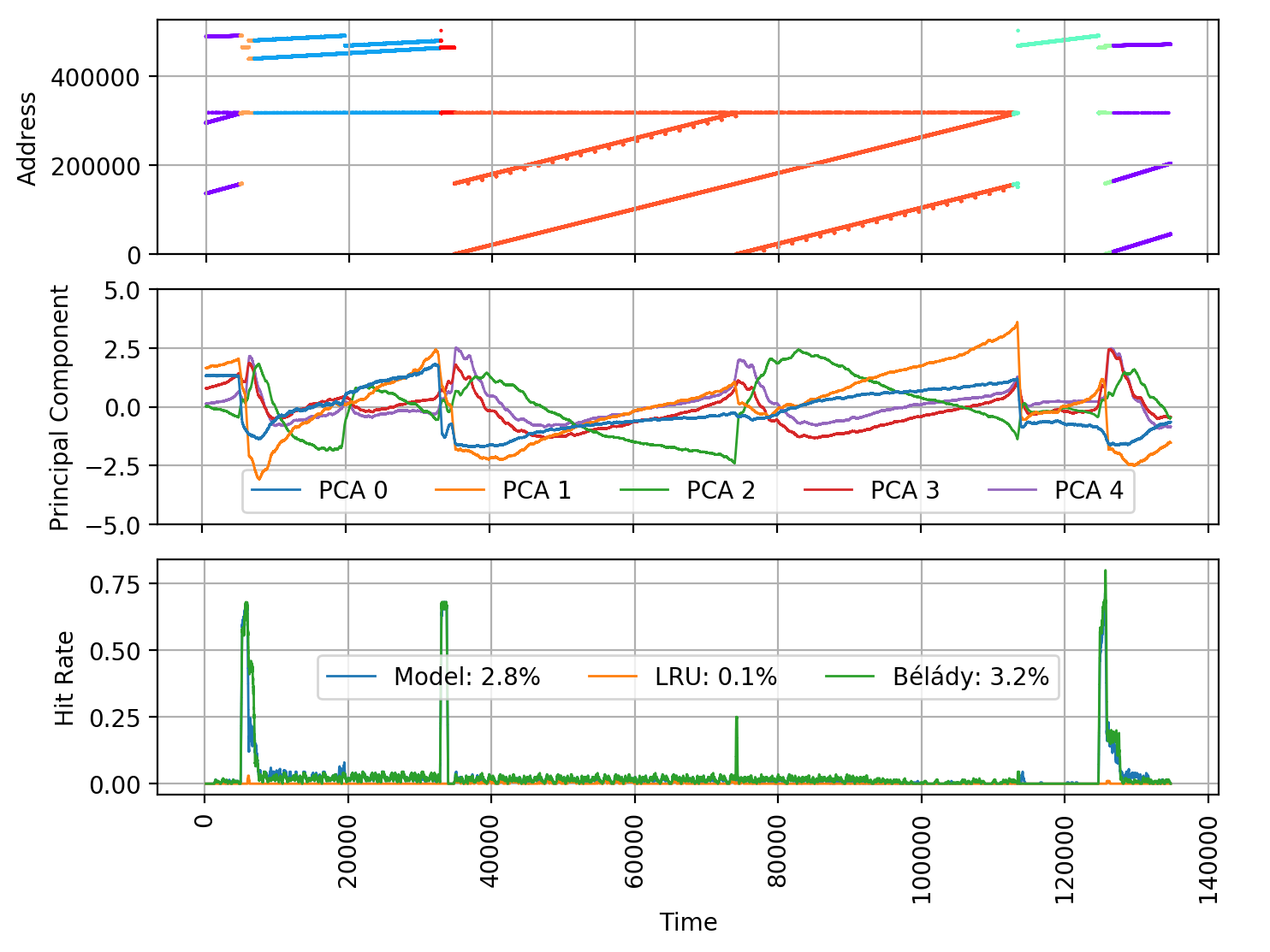}
    \caption{Original milc trace} \label{fig:pca_mlic_original} \end{subfigure}
        \begin{subfigure}{0.33\textwidth} \centering
        \includegraphics[width=0.85\linewidth]{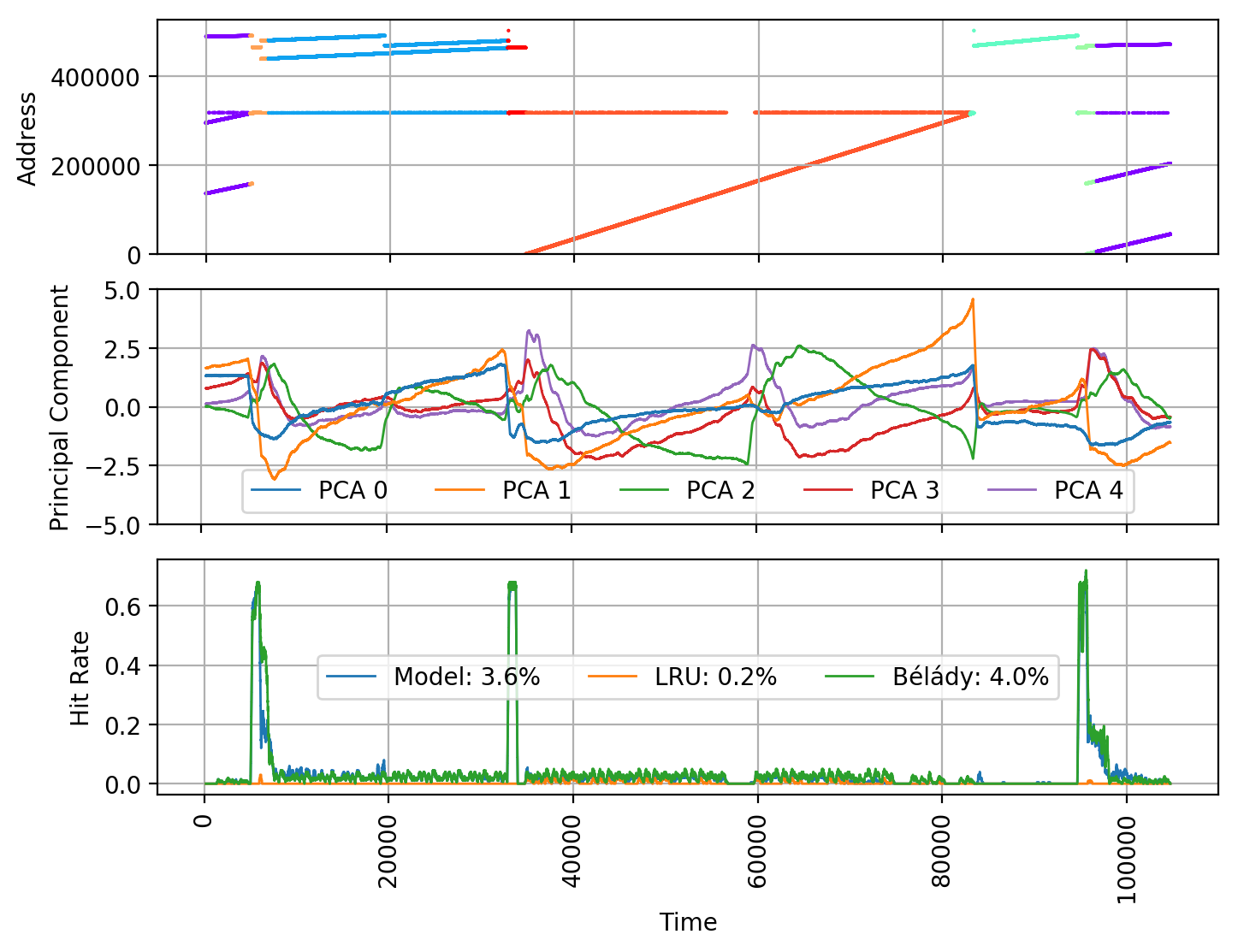}
        \caption{Milc after removing two streams } \label{fig:pca_mlic_22_25} \end{subfigure}
    \begin{subfigure}{0.33\textwidth} \centering
    \includegraphics[width=0.85\linewidth]{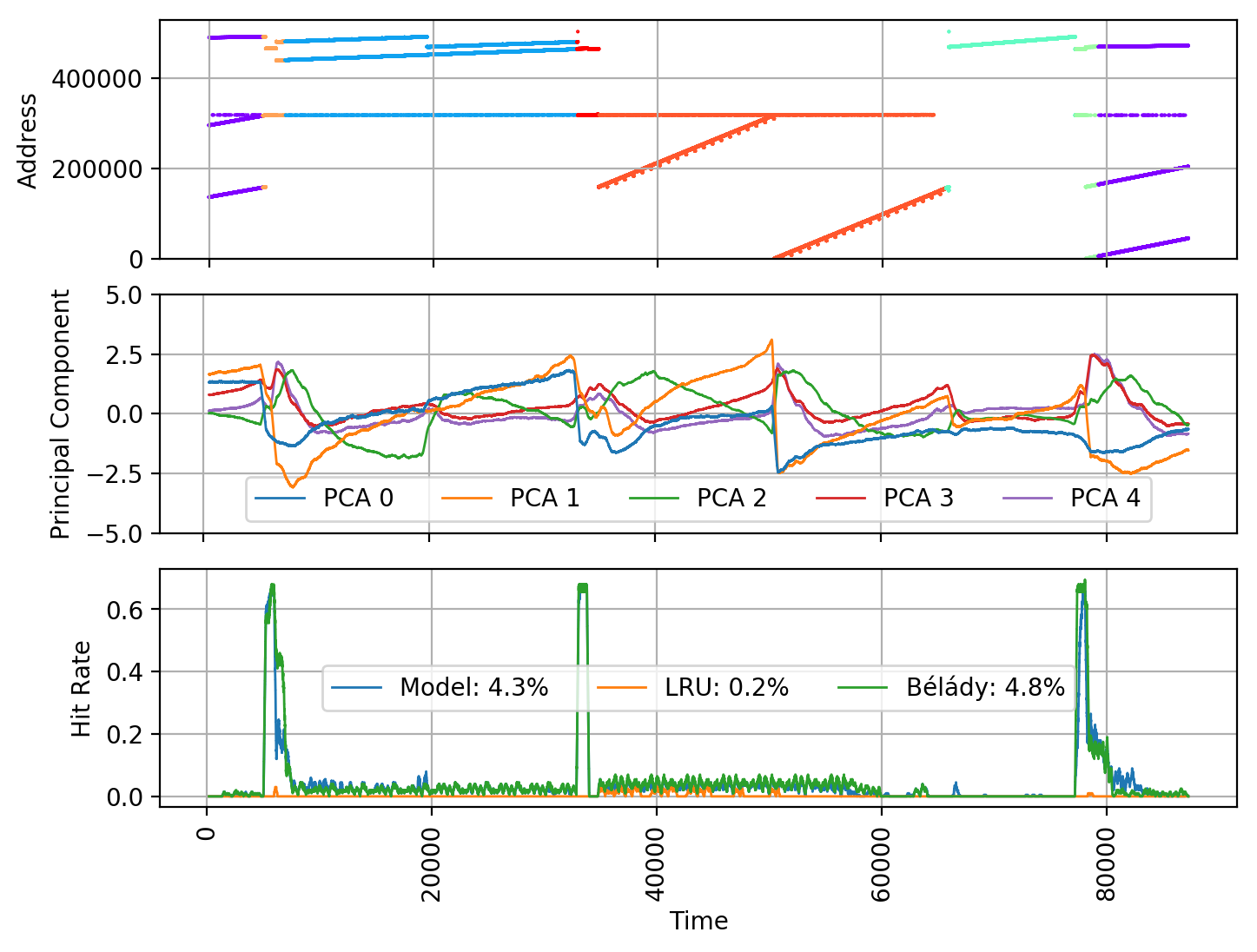}
    \caption{Milc after removing long stream} \label{fig:pca_mlic_24}
    \end{subfigure}
    \caption{
        The \textsc{milc} program has a phase with
        4 different streams: in orange \textbf{(a)}. The sign switch of the
        principal component 2 (green) co-occurs with the switch of the two
        smaller streams. However, removing them in \textbf{(b)} does not change
        the flip. Indeed, the PCs' magnitude barely changed (mean abs. difference: 0.075).
        When the long stream is deleted \textbf{(c)}, the green PCA2 changes
        to show less variability (mean abs. diff: 0.571). Additionally, the
        orange PCA1 switches sign.
        As the streams access the same memory, their
        differences has to come from the program counters. And indeed, each
        stream can be traced back to different program counters accessing the
        same memory.  }
    \label{fig:pca_milc}
\end{figure*}
}

\fakeparagraph{A simple baseline for caching based on simple statistics}
As the address embedding encodes
trace-specific information, a natural explanation of the model’s behavior would be that it memorizes which addresses to cache based on the current execution state.
To test this hypothesis, we construct a policy based on a simple lookup table and compare the model's performance to it. We also include the optimal \Belady policy as reference.
If a lookup policy performs similarly to the model, we would conclude that no high-level reasoning is needed.

\emph{Experiment}: We construct a simple caching policy based on
how often each address occurs per "phase"-type. A phase is a reoccurring
pattern within the execution. We isolate these phases from the trace using a hierarchical clustering based on statistics of
the program counter and reuse distance (i.e., the time until an address is used again; see Appendix \ref{sec:phases} for details).
The lookup table policy always evicts the least-used addresses given the current phase.
We then compare the hit rates between the simple lookup table, the \Belady's policy, and the model.

\emph{Result}: For most of the benchmarks, the simple policy
using the frequency of access alone is insufficient to achieve similar
performance to the model (see Table \ref{tab:phases}).  Therefore, the model must use a representation beyond these simple statistics.

\fakeparagraph{Inter-phase variability}
As we invalidated the lookup table hypothesis, we can reasonably conclude that the model is capable of more complex reasoning than a simple lookup table of frequency statistics and phases. However, what is the specific difference? Has the model a finer resolution than phases?
Which additional information does the model encode that is not included in simple statistics?

\emph{Experiment}:
To answer these questions, we inspect the model's internal state
by applying PCA to the LSTM's hidden state.
We use the hidden state as it is central to the model's behavior.
Any variation in model output is grounded either in the embeddings
or the hidden state, as they represent the attention queries, keys, and values.
PCA is the natural choice to describe the main directions of variance and
can be used without any labels. Generally, the lack of labels is
also the largest drawback of PCA, as the PCs do not have to correspond to
semantically meaningful concepts.
By looking at the top-5 principal components, we can
understand whether the model's activation varies within a phase or not.

\emph{Result}: As seen in Figure \ref{fig:pca_bzip_original}, the
principal components change within each phase in \textsc{bzip}.  While the
first principal component is strongly correlated
with the two main phases (r=0.89 and -0.83), other principal
components vary within each phase. Some of these changes seem to be
correlated with semantically meaningful changes. For example, a jump in four out
of five top principal components just before the end of the red phase in Figure
\ref{fig:pca_bzip_original} may indicate that the jump encodes the end of the
red phase -- a reasonable behaviour for models since some addresses are not
needed in the next phase.  On other programs, we can see similar behaviors. For
example, \textsc{leslie3d} also exhibits strong inter-phase variations (see PC1
in Figure \ref{fig:pca_activation_phases_leslie3d}).
These results indicate that the hidden state of the model has no simple interpretation and varies within each phase. This rules out simple explanations that would allow an engineer to reason about the model in terms of phases and ignore individual accesses.

\FigModificationsBzip

\fakeparagraph{Modifications of Streams}
The previous experiment suggests that the model encodes phases
and also variations within a phase.
This calls to analyze which patterns cause the variations.
In particular, we test what happens if a \emph{stream}
(a sequential access pattern) is removed.
A stream corresponds to a
linearly increasing memory addresses that are accessed in sequence, e.g by looping over an array.
An example can be seen in the lower lines in Figure \ref{fig:hit_miss}.
Both \emph{phases} and \emph{streams} are more abstract, or high-level, concepts than raw individual memory addresses. The question is: Is the model aware of these concepts?

\emph{Experiment}: We remove streams and see how the model behavior changes.
If the model encodes streams, we would expect to see changes in the hidden
activations.

\emph{Result (\bzip)}: Figure
\ref{fig:pca_bzip_removed_lower} shows the results of removing the streams on
\textsc{bzip}.  Removing the streams erases almost all variability in
principal components. Before erasing, PC0, PC2, PC3, PC4 make a distinctive jump
before the end of the red phase (see Figure \ref{fig:pca_bzip_original}).  Once
the streams are erased, the jumps are gone.  Therefore, the model relies on
the streams to encode the near ending of the phase.

\emph{Result (\textsc{milc})}: The milc program has a phase with
4 different streams (orange in Figure \ref{fig:pca_mlic_original}).  The
principal component 2 (green) switch its sign when the two smaller orange streams
switch. One could assume that as PC2 correlates with the streams it also
encodes information about them. However, this is not the case.  When the streams
are removed PC2 bearly changes (see Figure \ref{fig:pca_mlic_22_25}).  In
contrast, if the larger stream is removed, than PC2 does change (see Figure
\ref{fig:pca_mlic_24}).


%
%
%

\section{Conclusion}

We have provided a first set of analyses towards the interpretability of a caching model.
Specifically, we examined the level of abstraction at which the caching model
operates. We learned that while the address embedding memorizes trace-specific
information,
the model learns higher-level behaviour and can represent different phases
and their transitions.
It turned out that the concept of streams is particular relevant for encoding these transitions.
The conducted analysis shed some light into the typical behaviour of the model and the benefit of using concept-based explanations in caching models.
Nevertheless, further analysis is required
to close the gap between the way humans reason about programs and the caching model's low-level inputs.
An interesting research direction for future work would be to create a model that combines the two worlds of low-level machine behavior and high-level information, e.g., memory addresses or machine code with a summary of the source code.

\newpage

\bibliography{bibliography}

\begin{thebibliography}{14}
\providecommand{\natexlab}[1]{#1}
\providecommand{\url}[1]{\texttt{#1}}
\expandafter\ifx\csname urlstyle\endcsname\relax
  \providecommand{\doi}[1]{doi: #1}\else
  \providecommand{\doi}{doi: \begingroup \urlstyle{rm}\Url}\fi

\bibitem[{Bélády}(1966)]{belady1966}
L.~A. {Bélády}.
\newblock A study of replacement algorithms for a virtual-storage computer.
\newblock \emph{IBM Systems Journal}, 5\penalty0 (2):\penalty0 78--101, 1966.
\newblock \doi{10.1147/sj.52.0078}.

\bibitem[Cen et~al.(2020)Cen, Marcus, Mao, Gottschlich, Alizadeh, and
  Kraska]{lujing2020}
L.~Cen, R.~Marcus, H.~Mao, J.~Gottschlich, M.~Alizadeh, and T.~Kraska.
\newblock Learned garbage collection.
\newblock In \emph{Proceedings of the 4th ACM SIGPLAN International Workshop on
  Machine Learning and Programming Languages}, pages 38–--44, New York, NY,
  USA, 2020. Association for Computing Machinery.

\bibitem[Cohn and Singh(1997)]{lifetime_nips_1996}
D.~A. Cohn and S.~P. Singh.
\newblock Predicting lifetimes in dynamically allocated memory.
\newblock In M.~C. Mozer, M.~I. Jordan, and T.~Petsche, editors, \emph{Advances
  in Neural Information Processing Systems 9}, pages 939--945. MIT Press, 1997.

\bibitem[Hashemi et~al.(2018)Hashemi, Swersky, Smith, Ayers, Litz, Chang,
  Kozyrakis, and Ranganathan]{hashemi2018}
M.~Hashemi, K.~Swersky, J.~Smith, G.~Ayers, H.~Litz, J.~Chang, C.~Kozyrakis,
  and P.~Ranganathan.
\newblock Learning memory access patterns.
\newblock In \emph{Proceedings of the 35th International Conference on Machine
  Learning}, volume~80 of \emph{Proceedings of Machine Learning Research},
  pages 1919--1928, Stockholmsmässan, Stockholm Sweden, 10--15 Jul 2018. PMLR.

\bibitem[Henning(2006)]{henning2006spec}
J.~L. Henning.
\newblock Spec cpu2006 benchmark descriptions.
\newblock \emph{SIGARCH Comput. Archit. News}, 34\penalty0 (4):\penalty0
  1–17, Sept. 2006.
\newblock ISSN 0163-5964.
\newblock \doi{10.1145/1186736.1186737}.
\newblock URL \url{https://doi.org/10.1145/1186736.1186737}.

\bibitem[Liu et~al.(2020)Liu, Hashemi, Swersky, Ranganathan, and
  Ahn]{liu2020imitation}
E.~Liu, M.~Hashemi, K.~Swersky, P.~Ranganathan, and J.~Ahn.
\newblock An imitation learning approach for cache replacement.
\newblock In \emph{International Conference on Machine Learning}, pages
  6237--6247. PMLR, 2020.

\bibitem[Maas(2020)]{maas2020_ml_taxonomy}
M.~Maas.
\newblock A taxonomy of ml for systems problems.
\newblock \emph{IEEE Micro}, 40\penalty0 (5):\penalty0 8--16, 2020.

\bibitem[Maas et~al.(2020)Maas, Andersen, Isard, Javanmard, McKinley, and
  Raffel]{maas2020}
M.~Maas, D.~G. Andersen, M.~Isard, M.~M. Javanmard, K.~S. McKinley, and
  C.~Raffel.
\newblock Learning-based memory allocation for c++ server workloads.
\newblock In \emph{Proceedings of the Twenty-Fifth International Conference on
  Architectural Support for Programming Languages and Operating Systems},
  ASPLOS ’20, pages 541--556, New York, NY, USA, 2020. Association for
  Computing Machinery.

\bibitem[Mao et~al.(2019)Mao, Schwarzkopf, Venkatakrishnan, Meng, and
  Alizadeh]{mao2019}
H.~Mao, M.~Schwarzkopf, S.~B. Venkatakrishnan, Z.~Meng, and M.~Alizadeh.
\newblock Learning scheduling algorithms for data processing clusters.
\newblock In \emph{Proceedings of the ACM Special Interest Group on Data
  Communication}, SIGCOMM '19, pages 270–--288, New York, NY, USA, 2019.
  Association for Computing Machinery.

\bibitem[Shen et~al.(2018)Shen, Ling, Zhang, and Shi]{shen2018reusevectors}
S.~Shen, M.~Ling, Y.~Zhang, and L.~Shi.
\newblock Detecting the phase behavior on cache performance using the reuse
  distance vectors.
\newblock \emph{Journal of Systems Architecture}, 90:\penalty0 85 -- 93, 2018.
\newblock ISSN 1383-7621.
\newblock \doi{https://doi.org/10.1016/j.sysarc.2018.09.001}.
\newblock URL
  \url{http://www.sciencedirect.com/science/article/pii/S1383762118302248}.

\bibitem[Sherwood et~al.(2002)Sherwood, Perelman, Hamerly, and
  Calder]{sherwood2002simpoints}
T.~Sherwood, E.~Perelman, G.~Hamerly, and B.~Calder.
\newblock Automatically characterizing large scale program behavior.
\newblock In \emph{Proceedings of the 10th International Conference on
  Architectural Support for Programming Languages and Operating Systems},
  ASPLOS X, page 45–57, New York, NY, USA, 2002. Association for Computing
  Machinery.
\newblock ISBN 1581135742.
\newblock \doi{10.1145/605397.605403}.
\newblock URL \url{https://doi.org/10.1145/605397.605403}.

\bibitem[Song et~al.(2020)Song, Berger, Li, and Lloyd]{song2020}
Z.~Song, D.~S. Berger, K.~Li, and W.~Lloyd.
\newblock Learning relaxed belady for content distribution network caching.
\newblock In \emph{17th {USENIX} Symposium on Networked Systems Design and
  Implementation ({NSDI} 20)}, pages 529--544, Santa Clara, CA, 2020. {USENIX}
  Association.

\bibitem[Zhou and Maas(2021)]{zhou2021}
G.~Zhou and M.~Maas.
\newblock Learning on distributed traces for data center storage systems.
\newblock In A.~Smola, A.~Dimakis, and I.~Stoica, editors, \emph{Proceedings of
  Machine Learning and Systems}, volume~3, pages 350--364, 2021.
\newblock URL
  \url{https://proceedings.mlsys.org/paper/2021/file/82161242827b703e6acf9c726942a1e4-Paper.pdf}.

\bibitem[Zhou et~al.(2001)Zhou, Philbin, and Li]{cachesurvey2001}
Y.~Zhou, J.~Philbin, and K.~Li.
\newblock The multi-queue replacement algorithm for second level buffer caches.
\newblock In \emph{Proceedings of the General Track: 2001 USENIX Annual
  Technical Conference}, page 91–104, USA, 2001. USENIX Association.
\newblock ISBN 188044609X.

\end{thebibliography}


\newpage

\appendix

\section{Extraction of Program Phases}
\label{sec:phases}

Many execution traces show recurring patterns.
Examples include \bzip, which has two distinct phases, or \leslie.
(see Figure \ref{fig:phase_discovery}).
These phases typically correspond to different parts
of code in a program that execute one after another, such as loading an input
file and performing an operation on it.


\textbf{Phase Finding} The goal of the phase finding algorithm is
to extract highly similar phases using the input features of the trace.
Existing methods \citep{shen2018reusevectors,sherwood2002simpoints} find phases
using k-means. We decided against using them. Besides having to select the
number of clusters, k-means also has the disadvantage that a data point can be
far away from the cluster center as long as it is the closest center, which
can lead to very high variability within a phase.
However, we require a phases to be very uniform. If we characterize the model using phases,
we do not want to have to account for inter-phase variability.

The phase finding algorithm performs clustering based on Reuse Distance Vectors
\citep{shen2018reusevectors}. Reuse Distance Vectors compute a histogram over the
distance between accesses to the same address.
The reuse distance is closely related to
cacheability. As an additional feature, we use the histograms of the delta of
program counters.
The delta in program counters captures program execution at a mechanical level
(e.g., 1 means that the program stepped to the next instruction, a short
positive jump indicates a branch, and a long jump in any direction indicates
a function call).
In Figure \ref{fig:phase_discovery}, we show the two histograms
used as features to the phase finding.

We rely on the assumption that a trace will remain in
each phase for a meaningful amount of time.
We therefore first split a trace into small slices and then merge
neighbouring slices if they are highly similar. As metric, we use the
L1-distance between the histograms of the reuse distance and the deltas of the
program counters computed per slice.  Once no more neighbouring slices are
merged together, we find globally similar phases using the same metric.
We select the hyperparameters conservatively to
ensure only highly similar execution parts are merged together.

\FigPhases

\FigModificationsMilc


\FigAllHitMiss

\FigPcaEmbeddingAppx

\begin{figure*}
    \centering
    \newcommand{\pcaActivation}[2]{%
        \begin{subfigure}[t]{0.32\textwidth}
        \begin{center}
            \includegraphics[width=0.9\textwidth]{./figures/pca_activation_phases/#1.png}
        \end{center}
        \caption{#2}
        \label{fig:pca_activation_phases_#1}
        \end{subfigure}}

    \pcaActivation{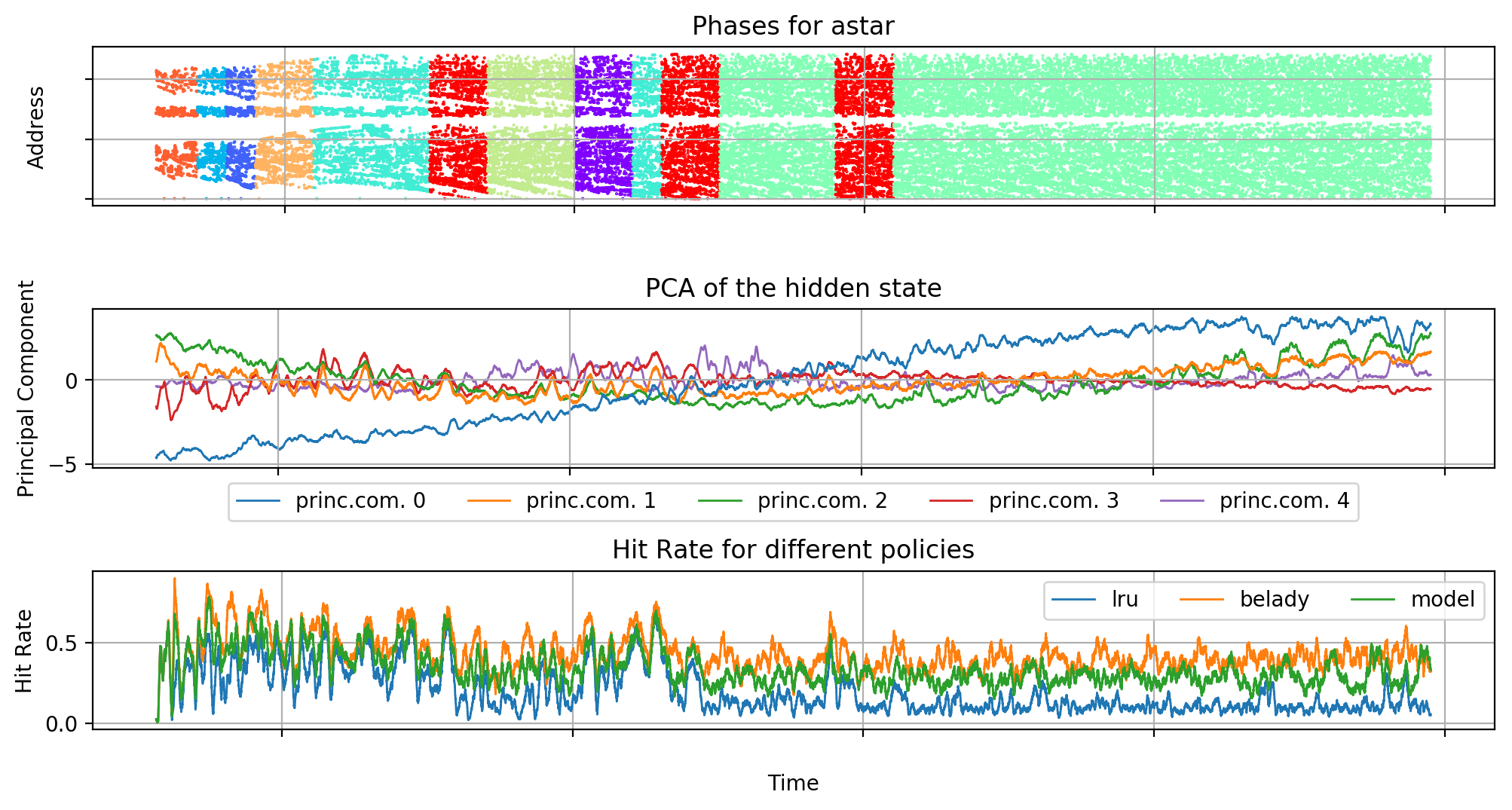}{\textsc{astar}}
    \pcaActivation{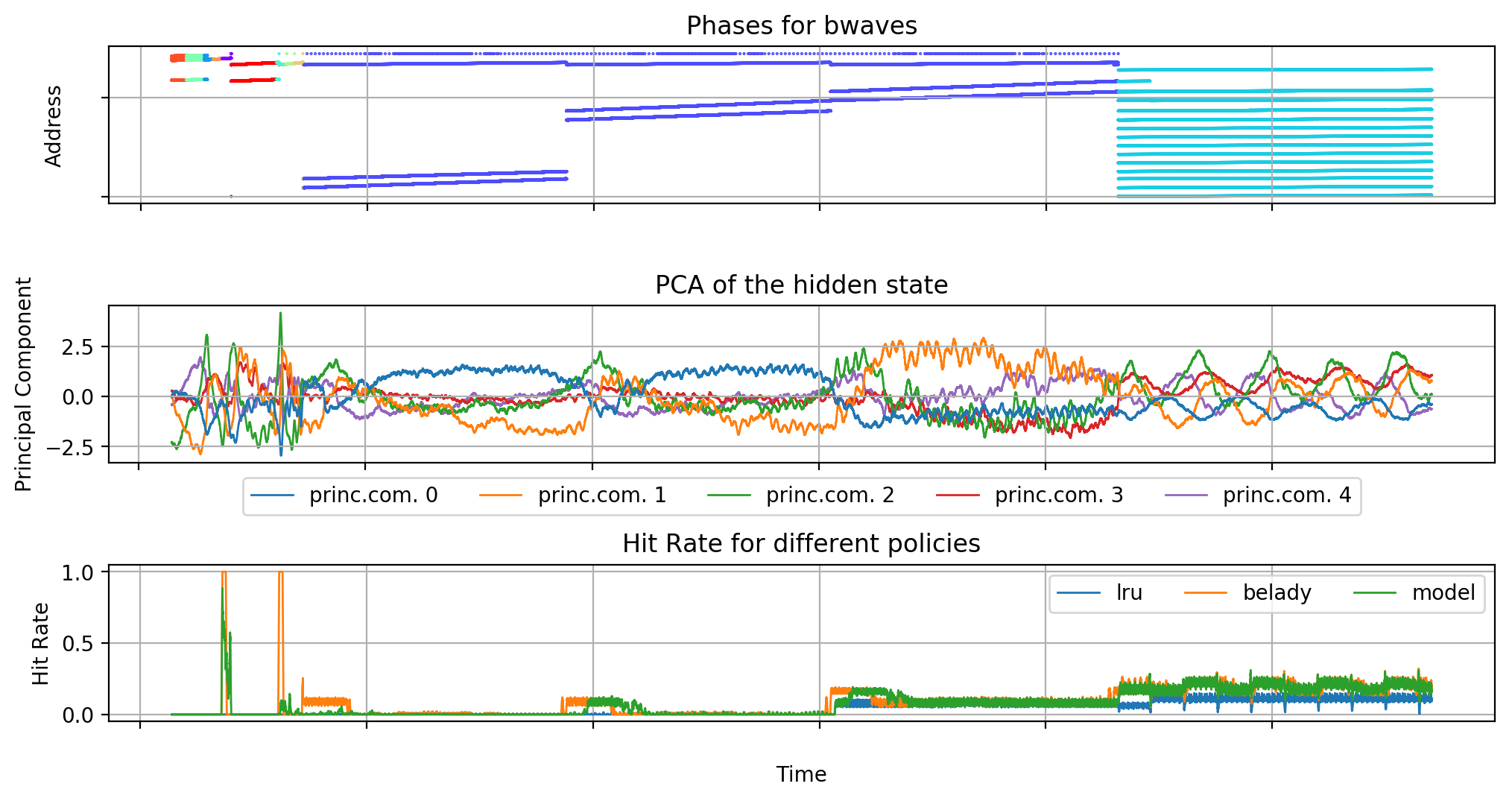}{\textsc{bwaves}}
    \pcaActivation{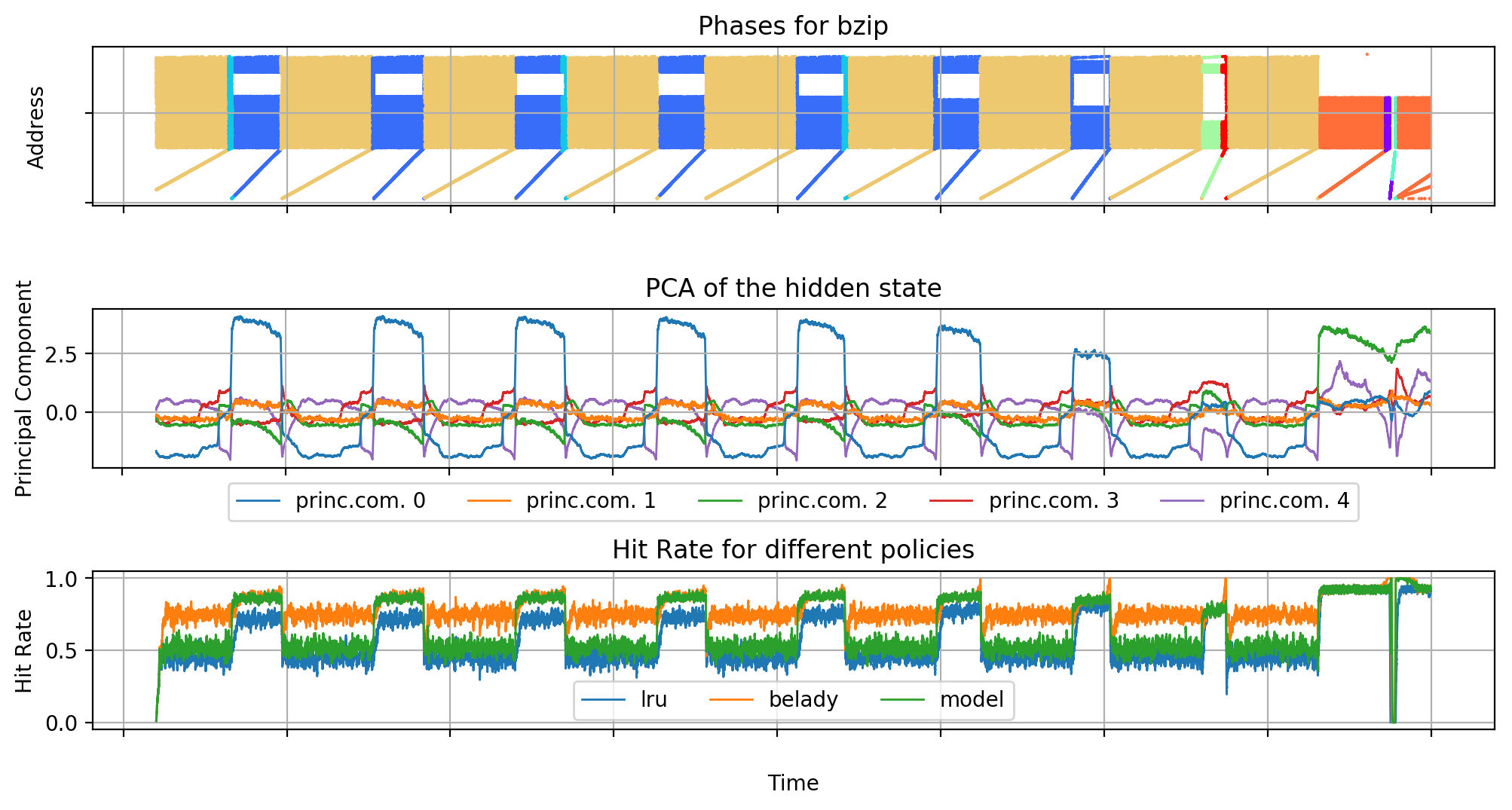}{\textsc{bzip}}
    \pcaActivation{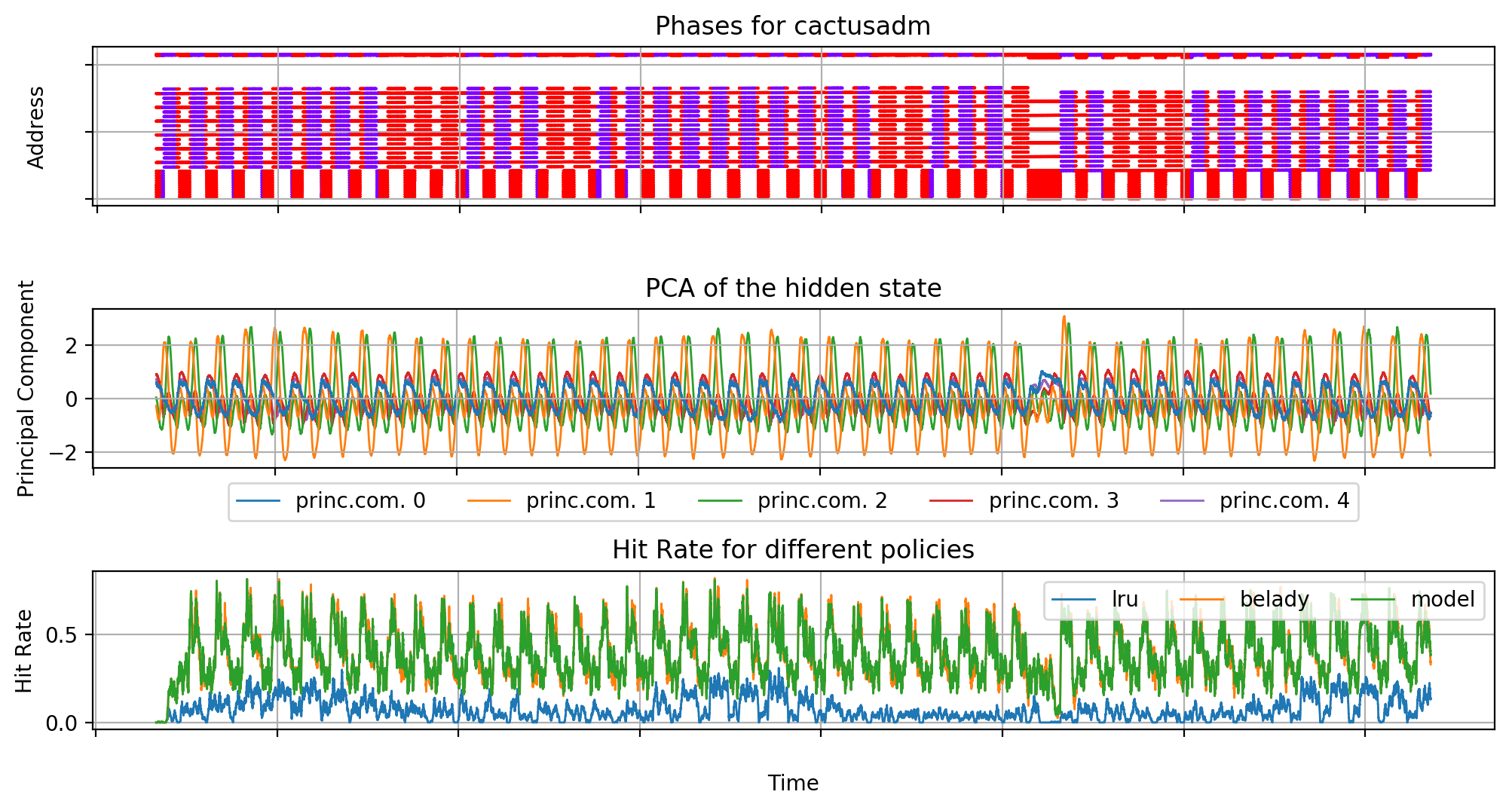}{\textsc{cactusadm}}
    \pcaActivation{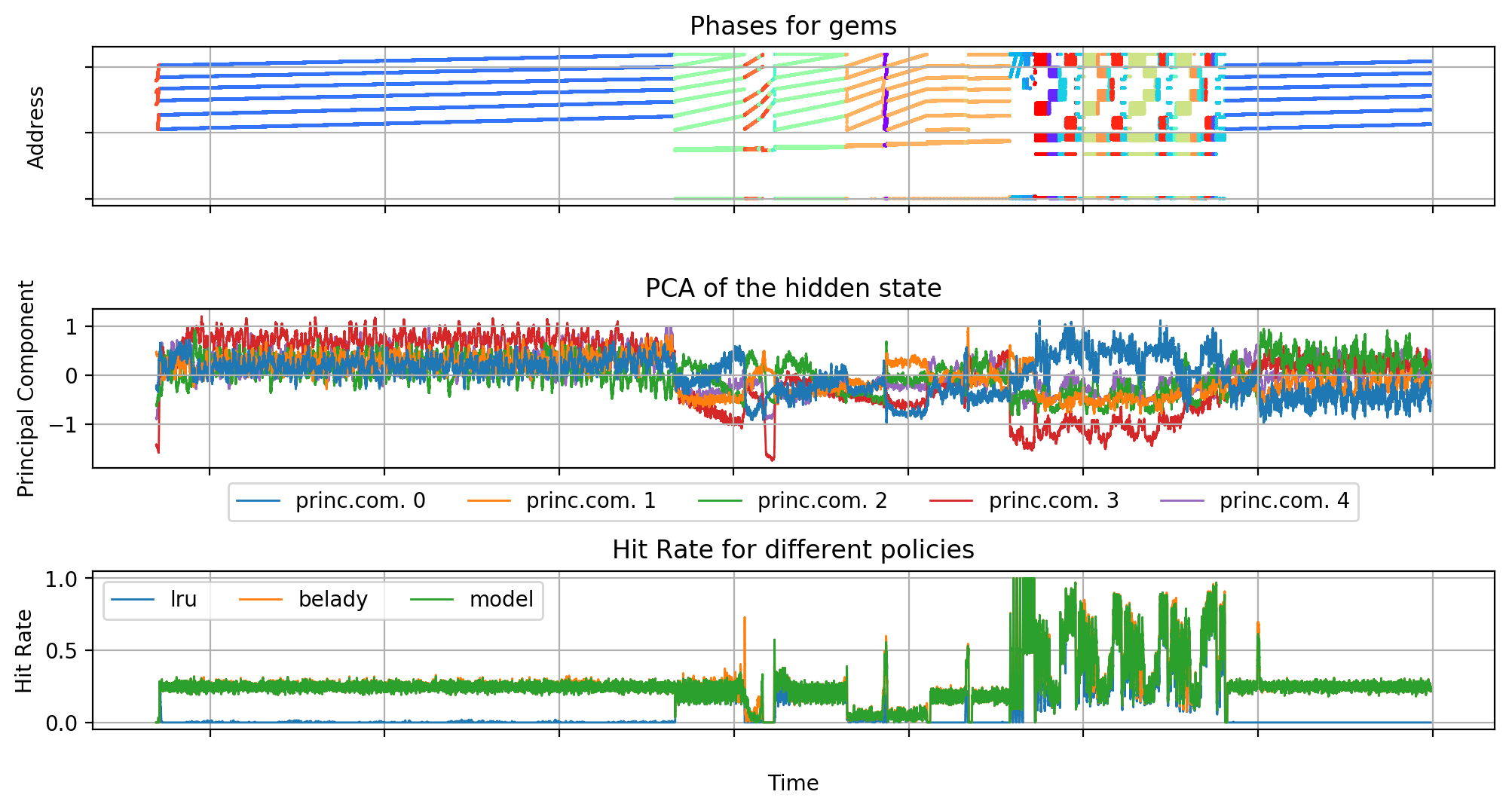}{\textsc{gems}}
    \pcaActivation{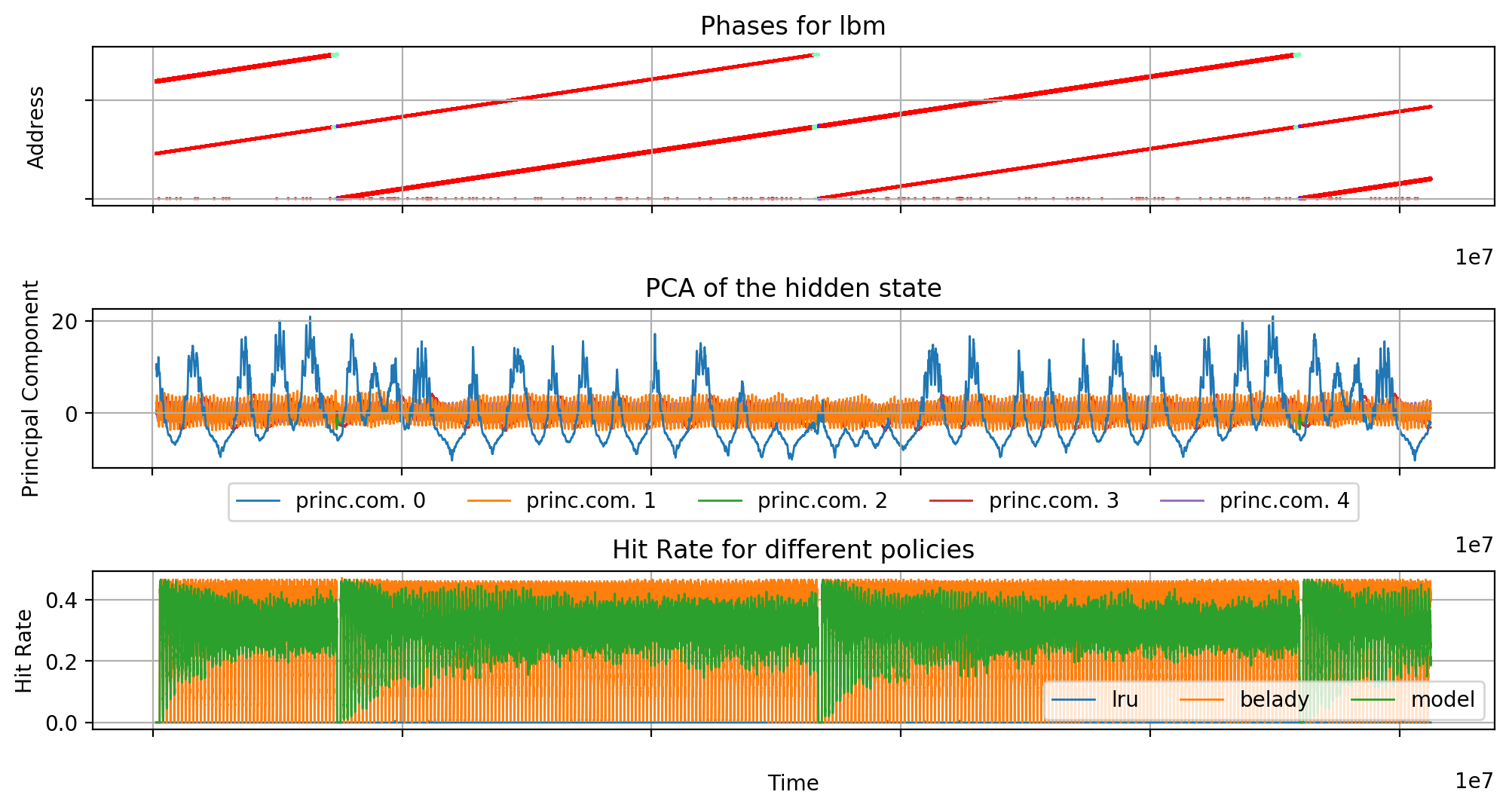}{\textsc{lbm}}
    \pcaActivation{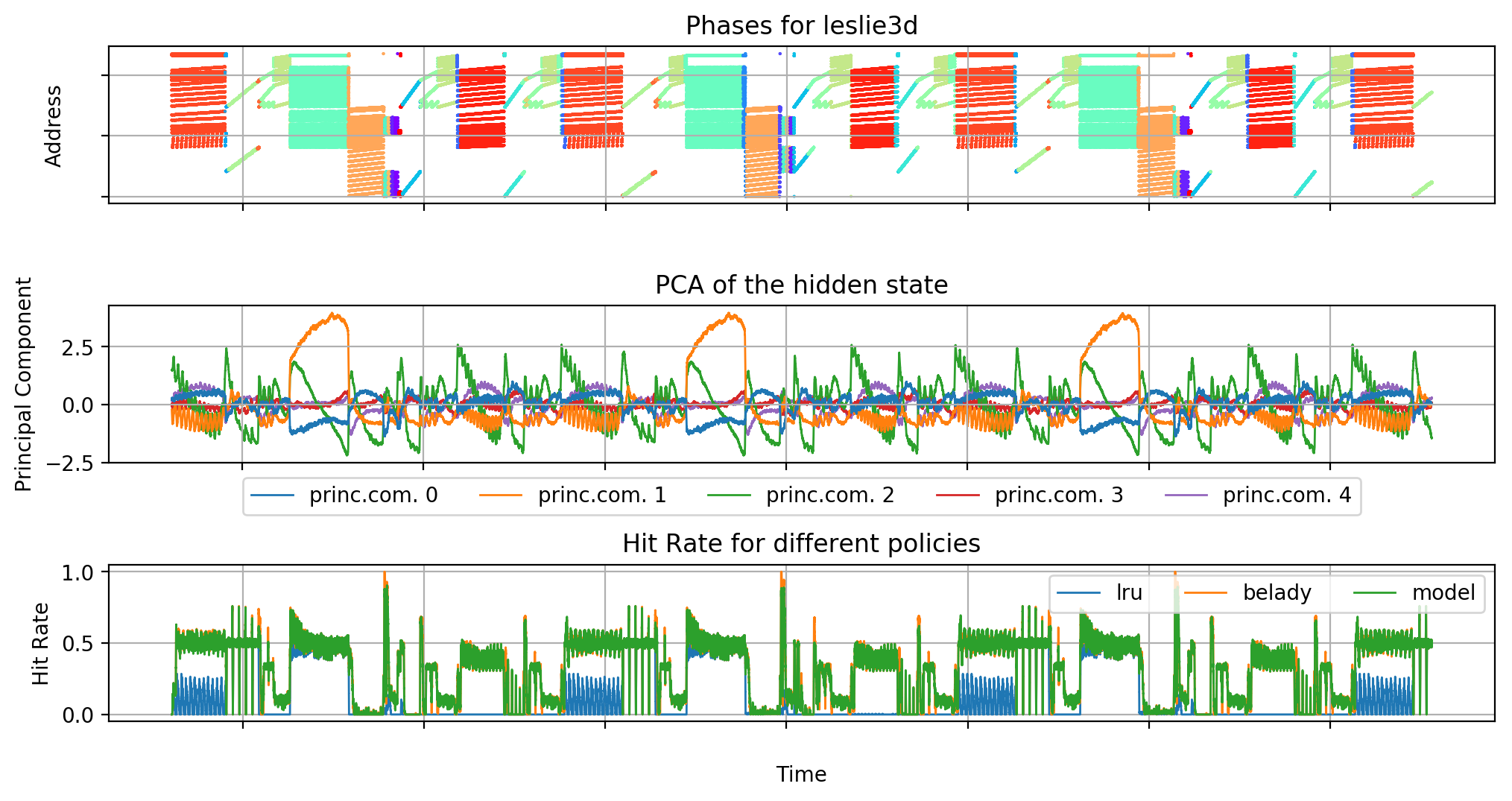}{\textsc{leslie3d}}
    \pcaActivation{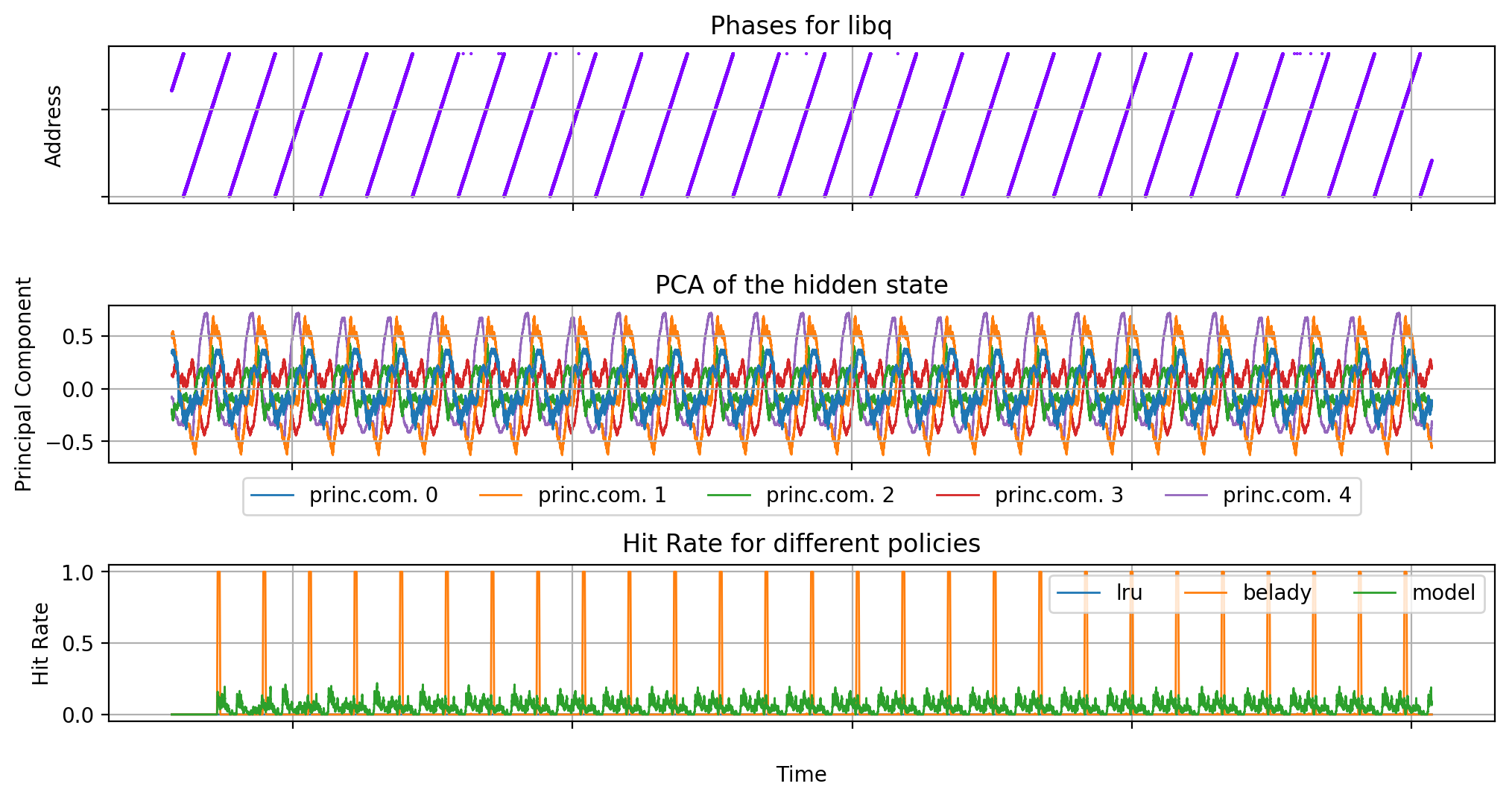}{\textsc{libq}}
    \pcaActivation{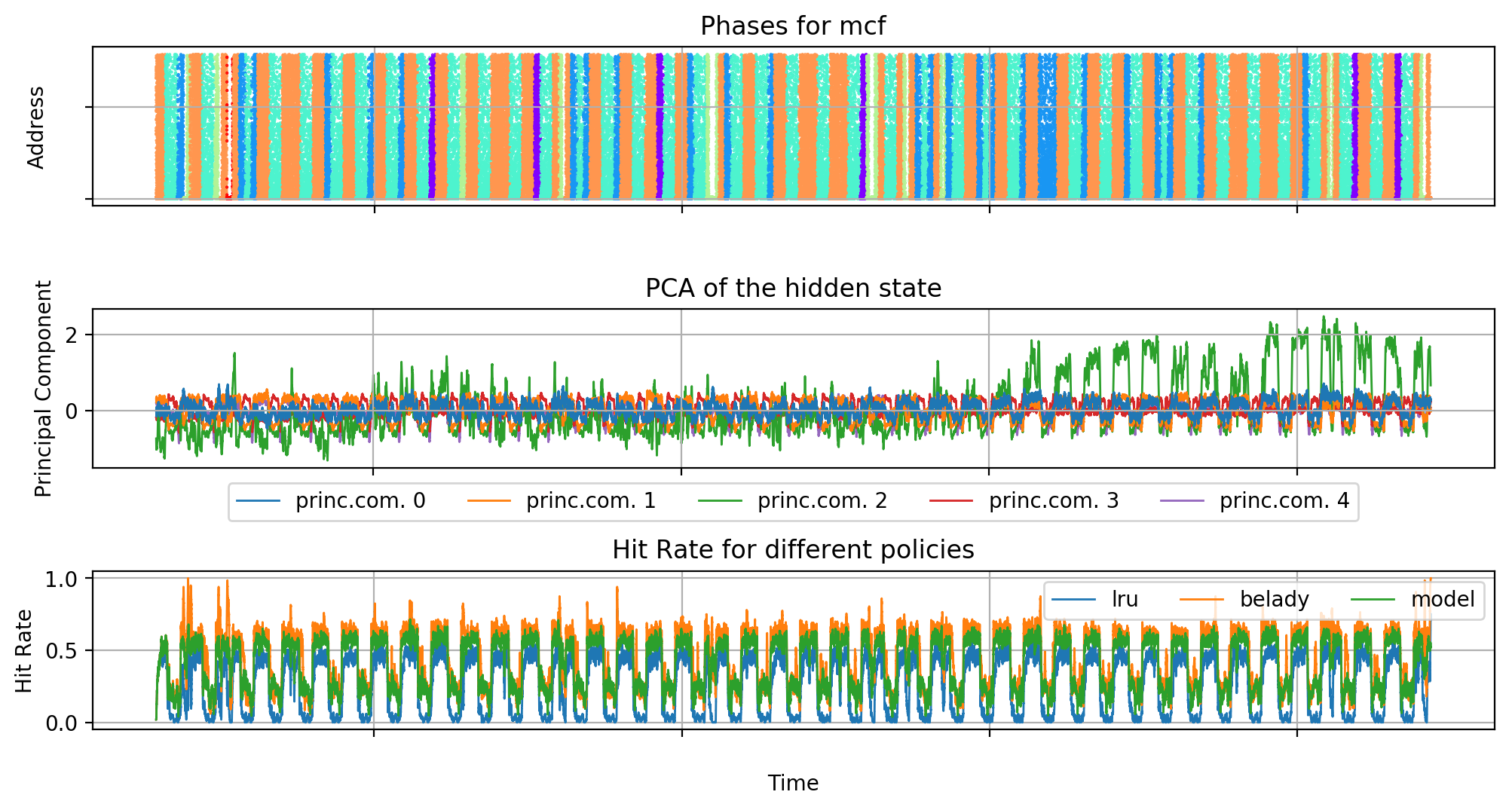}{\textsc{mcf}}
    \pcaActivation{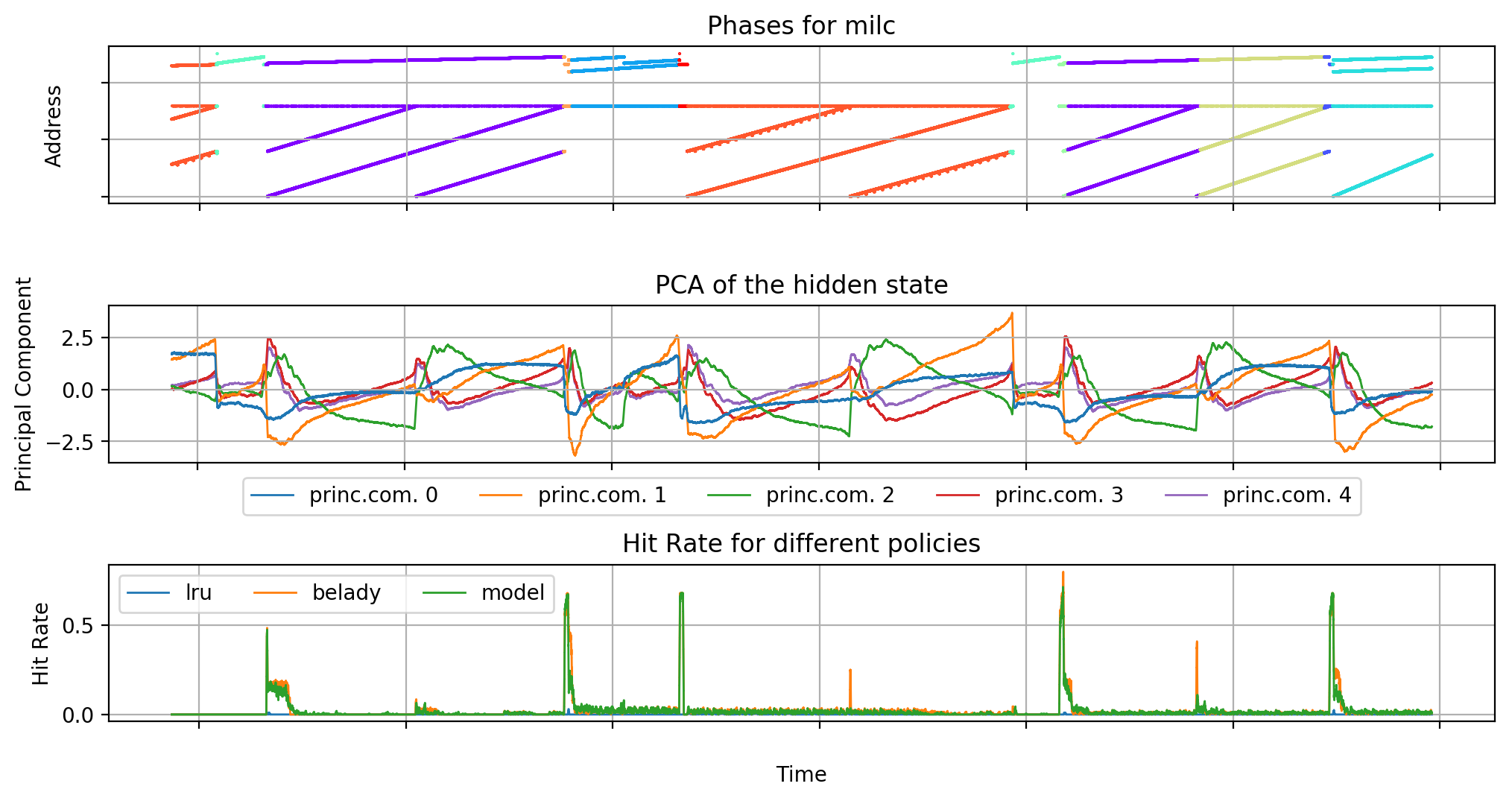}{\textsc{milc}}
    \pcaActivation{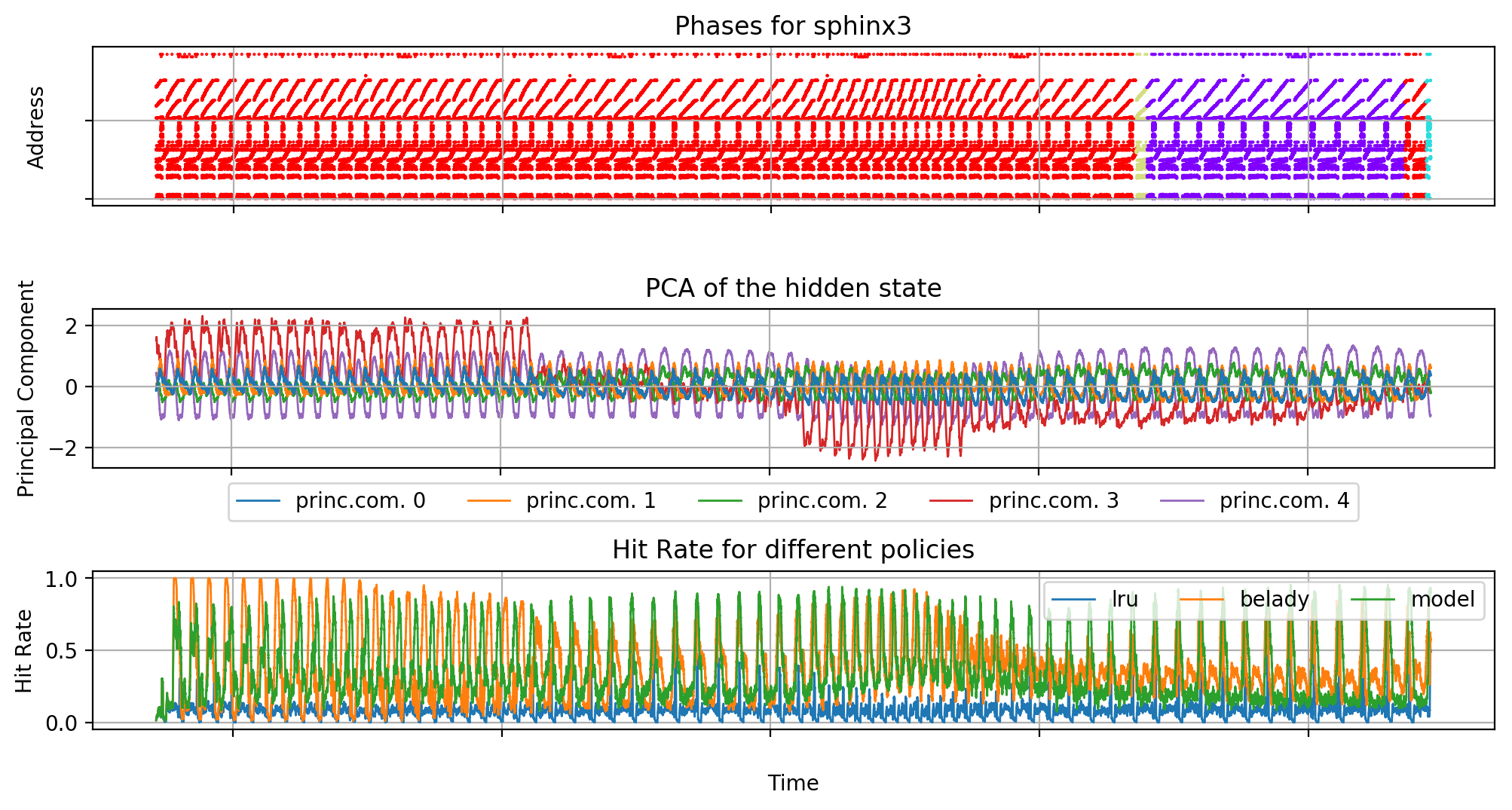}{\textsc{sphinx3}}
    \pcaActivation{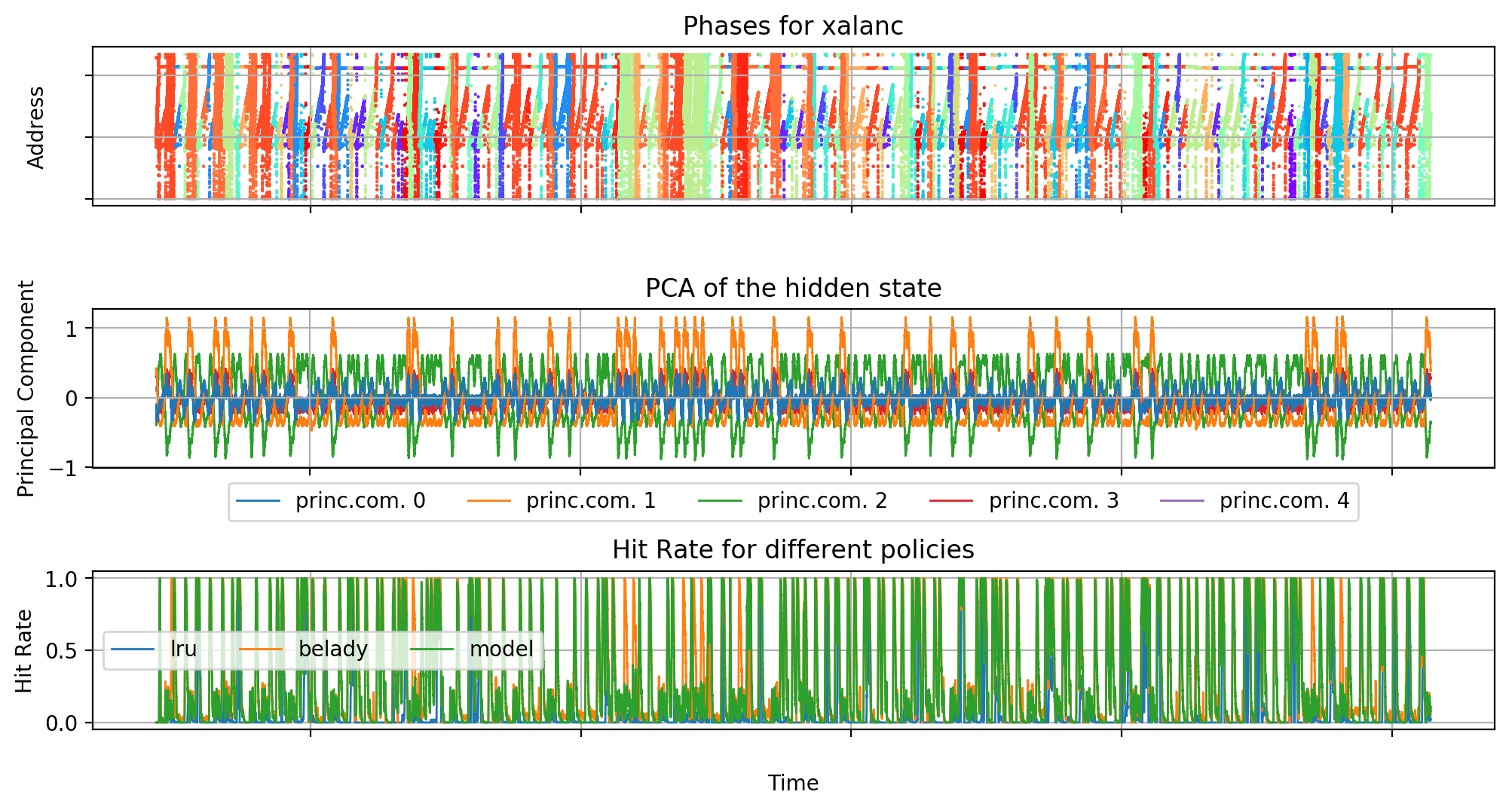}{\textsc{xalanc}}
    \caption{On the top, discovered phases. In the middle, the PCA of the hidden state.
    On the bottom, the hit rates of the learned policy, LRU and \Belady. }
    \label{fig:pca_activation_phases}
\end{figure*}

\end{document}